\documentclass[runningheads]{llncs}
\usepackage[export]{adjustbox}
\usepackage{graphicx}
\usepackage[percent]{overpic}
\usepackage{comment}
\usepackage{amsmath,amssymb} %
\usepackage{booktabs}
\usepackage{multirow}
\usepackage{gensymb}

\newcommand{\algname}{\mbox{Trajectron++}}
\newcommand{\emphalgname}{\emph{\algname}}

\usepackage[pagebackref=true,breaklinks=true,colorlinks,bookmarks=false]{hyperref}
\usepackage[noabbrev,capitalise]{cleveref}

\usepackage{url}

\newcommand\Tstrut{\rule{0pt}{2.6ex}}       %
\newcommand\Bstrut{\rule[-0.9ex]{0pt}{0pt}} %
\newcommand{\TBstrut}{\Tstrut\Bstrut} %

\begin{document}
\pagestyle{headings}
\mainmatter
\def\ECCVSubNumber{3094}  %

\title{
\algname{}: Dynamically-Feasible Trajectory Forecasting With Heterogeneous Data
}

%
%

%
%
\titlerunning{Trajectron++: Dynamically-Feasible Trajectory Forecasting}
\author{Tim Salzmann\thanks{Equal contribution.}\thanks{Work done as a visiting student in the Autonomous Systems Lab.}\inst{1} \and
Boris Ivanovic$^\star$\inst{1} \and %
Punarjay Chakravarty\inst{2} \and %
\\Marco Pavone\inst{1}} %
\authorrunning{T. Salzmann$^\star$, B. Ivanovic$^\star$, et al.}
\institute{Autonomous Systems Lab, Stanford University\\
\email{\{timsal, borisi, pavone\}@stanford.edu} 
\and 
Ford Greenfield Labs\\
\email{pchakra5@ford.com}}
\maketitle
\setcounter{footnote}{0}
\begin{abstract}
   Reasoning about human motion is an important prerequisite to safe and socially-aware robotic navigation. 
   As a result, multi-agent behavior prediction has become a core component of modern human-robot interactive systems, such as self-driving cars. 
   While there exist many methods for trajectory forecasting, most do not enforce dynamic constraints and do not account for environmental information (e.g., maps). 
   Towards this end, we present \emphalgname{}, a modular, graph-structured recurrent model that forecasts the trajectories of a general number of diverse agents while incorporating agent dynamics and heterogeneous data (e.g., semantic maps).
   \emphalgname{} is designed to be tightly integrated with robotic planning and control frameworks; for example, it can produce predictions that are optionally conditioned on ego-agent motion plans. 
   We demonstrate its performance on several challenging real-world trajectory forecasting datasets, outperforming a wide array of state-of-the-art deterministic and generative methods.

\end{abstract}

\section{Introduction}

\begin{figure}[t]
    \centering
    \includegraphics[trim={0 23em 0 17em},clip,width=\linewidth]{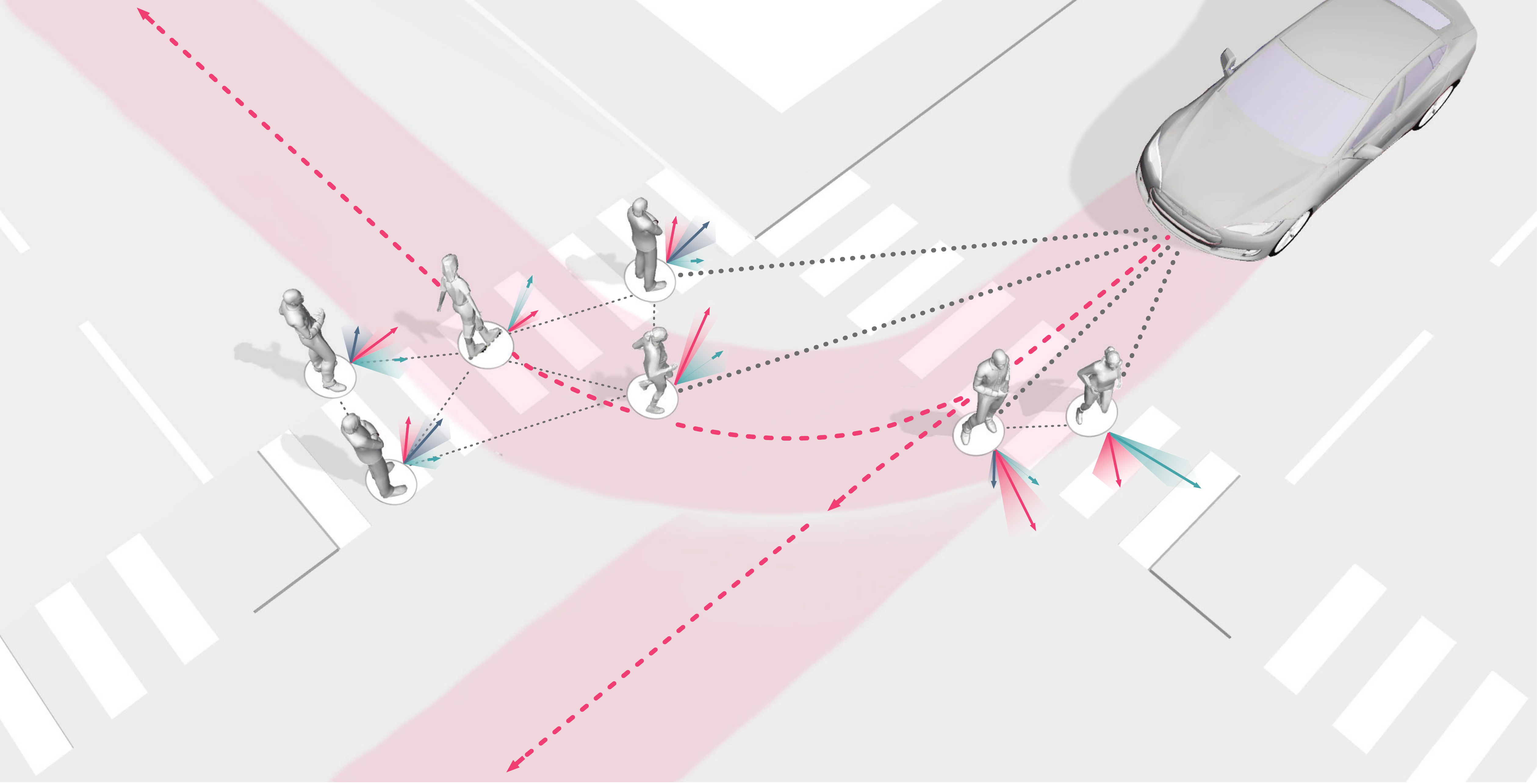}
    \caption{Exemplary road scene depicting pedestrians crossing a road in front of a vehicle which may continue straight or turn right. The graph representation of the scene is shown on the ground, where each agent and their interactions are represented as nodes and edges, visualized as white circles and dashed black lines, respectively. Arrows depict potential future agent velocities, with colors representing different high-level future behavior modes.}
    \label{fig:hero}
\end{figure}

Predicting the future behavior of humans is a necessary part of developing safe human-interactive autonomous systems. Humans can naturally navigate through many social interaction scenarios because they have an intrinsic ``theory of mind,'' which is the capacity to reason about other people's actions in terms of their mental states \cite{GweonSaxe2013}.
As a result, imbuing autonomous systems with this capability
could enable more informed decision making and proactive actions to be taken in the presence of other intelligent agents, e.g., in human-robot interaction scenarios. \cref{fig:hero} illustrates a scenario where predicting the intent of other agents may inform an autonomous vehicle's path planning and decision making. Indeed, multi-agent behavior prediction has already become a core component of modern robotic systems, especially in safety-critical applications like self-driving vehicles which are currently being tested in the real world and targeting widespread deployment in the near future \cite{WaymoSafety2018}.

There are many existing methods for multi-agent behavior prediction, ranging from deterministic regressors to generative, probabilistic models. However, many of them were developed without directly accounting for real-world robotic use cases; in particular, they ignore agents' dynamics constraints, the ego-agent's own motion (important to capture the interactive aspect in human-robot interaction), and 
a plethora of environmental information (e.g., camera images, lidar, maps) to which modern robotic systems have access. \cref{tab:related_work} provides a summary of recent state-of-the-art approaches and their consideration of such desiderata.

Accordingly, in this work we are interested in developing a multi-agent behavior prediction model that 
(1)~accounts for the dynamics of the agents, and in particular of ground vehicles~\cite{KongPfeiferEtAl2015,PadenCapEtAl2016};
(2)~produces predictions possibly conditioned on potential future robot trajectories, useful for intelligent planning taking into account human responses; and
(3)~provides a generally-applicable, open, and extensible approach which can effectively use heterogeneous data about the surrounding environment.
Importantly, making use of such data would allow for the incorporation of environmental information, e.g., maps, which would enable producing predictions that differ depending on the structure of the scene (e.g., interactions at an urban intersection are very different from those in an open sports field!).
One method that comes close is the Trajectron~\cite{IvanovicPavone2019}, a multi-agent behavior model which can handle a time-varying number of agents, accounts for multimodality in human behavior (i.e., the potential for many high-level futures), and maintains a sense of interpretability in its outputs.
However, the Trajectron only reasons about relatively simple vehicle models (i.e., cascaded integrators) and past trajectory data (i.e., no considerations are made for added environmental information, if available).

In this work we present \emphalgname{}, an open and extensible approach built upon the Trajectron \cite{IvanovicPavone2019} framework which produces dynamically-feasible trajectory forecasts from heterogeneous input data for multiple interacting agents of distinct semantic types.
Our key contributions are twofold: First, we show how to effectively incorporate high-dimensional data
through the lens of encoding semantic maps. Second, we propose a general method of incorporating dynamics constraints into learning-based methods for multi-agent trajectory forecasting.
\emphalgname{} is designed to be tightly integrated with downstream robotic modules,
with the ability to produce trajectories
that are optionally conditioned on future ego-agent motion plans.
We present experimental results on a variety of datasets, which collectively demonstrate that \emphalgname{} outperforms an extensive selection of state-of-the-art deterministic and generative trajectory prediction methods, in some cases achieving $60\%$ lower average prediction error.

\section{Related Work}

\begin{table}[t]
\fontsize{8}{8}\selectfont
\centering
\caption{A summary of recent state-of-the-art pedestrian (left) and vehicle (right) trajectory forecasting methods, indicating the desiderata addressed by each approach.}
\begin{tabular}{l|ccccc}
\toprule
\textbf{Method} & GNA & CD & HD & FCP & OS \TBstrut \\ \midrule
DESIRE \cite{LeeChoiEtAl2017} & \checkmark & & \checkmark & & \\
Trajectron \cite{IvanovicPavone2019} & \checkmark & & & & \checkmark\\ 
S-BiGAT \cite{KosarajuSadeghianEtAl2019} & \checkmark & & \checkmark & & \\
DRF-Net \cite{JainCasasEtAl2019} & \checkmark & & \checkmark & & \\
MATF \cite{ZhaoXuEtAl2019} & \checkmark & & \checkmark & & \checkmark \\
\midrule 
Our Work & \checkmark & \checkmark & \checkmark & \checkmark & \checkmark\\
\bottomrule
\end{tabular}
\begin{tabular}{l|ccccc}
\toprule
\textbf{Method} & GNA & CD & HD & FCP & OS \TBstrut \\ \midrule
IntentNet \cite{CasasLuoEtAl2018} & \checkmark & & \checkmark & & \\
PRECOG \cite{RhinehartMcAllisterEtAl2019} & & & \checkmark & \checkmark & \checkmark\\
MFP \cite{TangSalakhutdinov2019} & \checkmark & & \checkmark & & \checkmark \\
NMP \cite{ZengLuoEtAl2019} & \checkmark & & \checkmark & & \\
SpAGNN \cite{CasasGulinoEtAl2019} & \checkmark & & \checkmark & & \\
\midrule 
Our Work & \checkmark & \checkmark & \checkmark & \checkmark & \checkmark\\
\bottomrule
\end{tabular}

Legend: GNA = General Number of Agents, CD = Considers Dynamics, HD = Heterogeneous Data, FCP = Future-Conditional Predictions, OS = Open Source

\label{tab:related_work}
\end{table}

{\bf Deterministic Regressors.} Many earlier works in human trajectory forecasting were deterministic regression models. One of the earliest, the Social Forces model \cite{HelbingMolnar1995}, models humans as physical objects affected by Newtonian forces (e.g., with attractors at goals and repulsors at other agents). Since then, many approaches have been applied to the problem of trajectory forecasting, formulating it as a time-series regression problem and applying methods like Gaussian Process Regression (GPR) \cite{RasmussenWilliams2006,WangFleetEtAl2008}, Inverse Reinforcement Learning (IRL) \cite{LeeKitani2016}, and Recurrent Neural Networks (RNNs) \cite{AlahiGoelEtAl2016,MortonWheelerEtAl2017,VemulaMuellingEtAl2018} to good effect. An excellent review of such methods can be found in \cite{RudenkoPalmieriEtAl2019}.

{\bf Generative, Probabilistic Approaches.} Recently, generative approaches have emerged as state-of-the-art trajectory forecasting methods due to recent advancements in deep generative models \cite{SohnLeeEtAl2015,GoodfellowPouget-AbadieEtAl2014}. Notably, they have caused a shift from focusing on predicting the single best trajectory to producing a \textit{distribution} of potential future trajectories. This is advantageous in autonomous systems as full distribution information is more useful for downstream tasks, e.g., motion planning and decision making where information such as variance can be used to make safer decisions. Most works in this category use a deep recurrent backbone architecture with a latent variable model, such as a Conditional Variational Autoencoder (CVAE) \cite{SohnLeeEtAl2015}, to explicitly encode multimodality \cite{LeeChoiEtAl2017,IvanovicSchmerlingEtAl2018,DeoTrivedi2018,SadeghianLegrosEtAl2018,IvanovicPavone2019,RhinehartMcAllisterEtAl2019}, or a Generative Adversarial Network (GAN) \cite{GoodfellowPouget-AbadieEtAl2014} to implicitly do so \cite{GuptaJohnsonEtAl2018,SadeghianKosarajuEtAl2019,KosarajuSadeghianEtAl2019,ZhaoXuEtAl2019}. Common to both approach styles is the need to produce position distributions. GAN-based models can directly produce these and CVAE-based recurrent models usually rely on a bivariate Gaussian Mixture Model (GMM) to output position distributions. However, both of these output structures make it difficult to enforce dynamics constraints, e.g., non-holonomic constraints such as those arising from no side-slip conditions.
Of these, the Trajectron \cite{IvanovicPavone2019} and MATF \cite{ZhaoXuEtAl2019} are the best-performing CVAE-based and GAN-based models, respectively, on standard pedestrian trajectory forecasting benchmarks \cite{PellegriniEssEtAl2009,LernerChrysanthouEtAl2007}. 

{\bf Accounting for Dynamics and Heterogeneous Data.} There are few works that account for dynamics or make use of data modalities outside of prior trajectory information. This is mainly because standard trajectory forecasting benchmarks seldom include any other information, a fact that will surely change following the recent release of autonomous vehicle-based datasets with rich multi-sensor data \cite{waymo_open_dataset,CaesarBankitiEtAl2019,ChangLambertEtAl2019,lyft_dataset2019}.
As for dynamics, current methods almost exclusively reason about positional information. This does not capture dynamical constraints, however, which might lead to predictions in position space that are unrealizable by the underlying control variables (e.g., a car moving sideways).
\cref{tab:related_work} provides a detailed breakdown of recent state-of-the-art approaches and their consideration of these desiderata.

\section{Problem Formulation}
We aim to generate plausible trajectory distributions for a time-varying number $N(t)$ of interacting agents $A_1,...,A_{N(t)}$. Each agent $A_i$ has a semantic class $S_i$, e.g., Car, Bus, or Pedestrian. At time $t$, given 
the state $\mathbf{s} \in \mathbb{R}^D$ of each agent
and all of their histories for the previous $H$ timesteps, which we denote as $\mathbf{x}$, $\mathbf{x} = \mathbf{s}_{1,\dots,N(t)}^{(t - H : t)} \in \mathbb{R}^{(H + 1) \times N(t) \times D}$, as well as additional information available to each agent $I_{1,\dots,N(t)}^{(t)}$, we seek a distribution over all agents' future states for the next $T$ timesteps $\mathbf{y} = \mathbf{s}_{1,\dots,N(t)}^{(t + 1 : t + T)} \in \mathbb{R}^{T \times N(t) \times D}$, which we denote as $p(\mathbf{y} \mid \mathbf{x}, I)$.
We also assume that geometric semantic maps are available around $A_i$'s position, $M_i^{(t)} \in \mathbb{R}^{\left \lceil C/r \right \rceil \times \left \lceil C/r \right \rceil \times L}$, with context size $C \times C$, spatial resolution $r$, and $L$ semantic channels. Depending on the dataset, these maps can range in sophistication from simple obstacle occupancy grids to multiple layers of human-annotated semantic information (e.g., marking out sidewalks, road boundaries, and crosswalks). 

We also consider the setting where we condition on an ego-agent's future motion plan, for example when evaluating responses to a set of motion primitives. In this setting, we additionally assume that we know the ego-agent's future motion plan for the next $T$ timesteps, $\mathbf{y}_\text{R} = \mathbf{s}_\text{R}^{(t + 1 : t + T)}$.

\section{\algname{}} \label{sec:model}
\begin{figure}[t]
    \centering
    \raisebox{-0.5\height}{\includegraphics[width=0.4\linewidth]{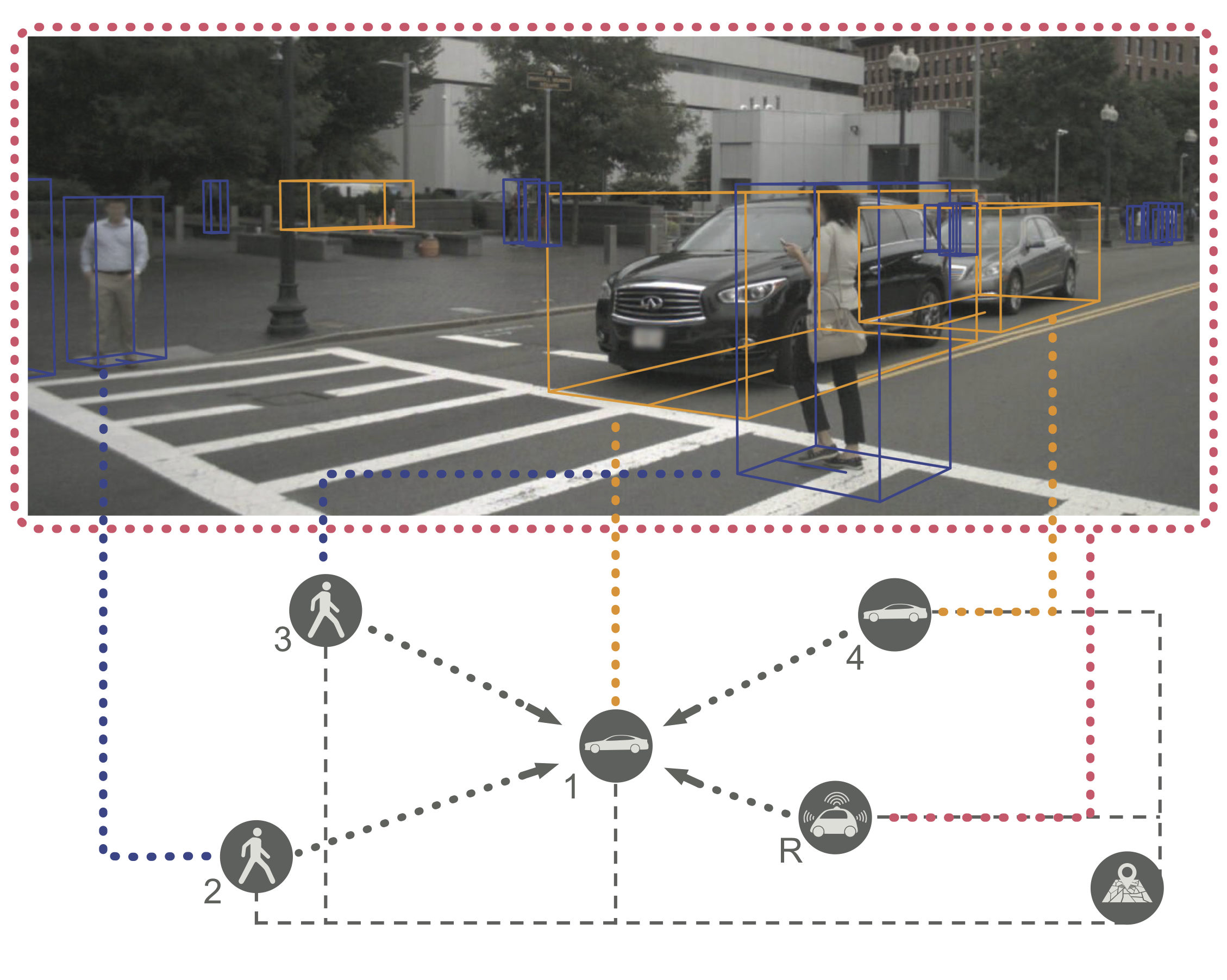}}
    \raisebox{-0.5\height}{\includegraphics[width=0.5\linewidth]{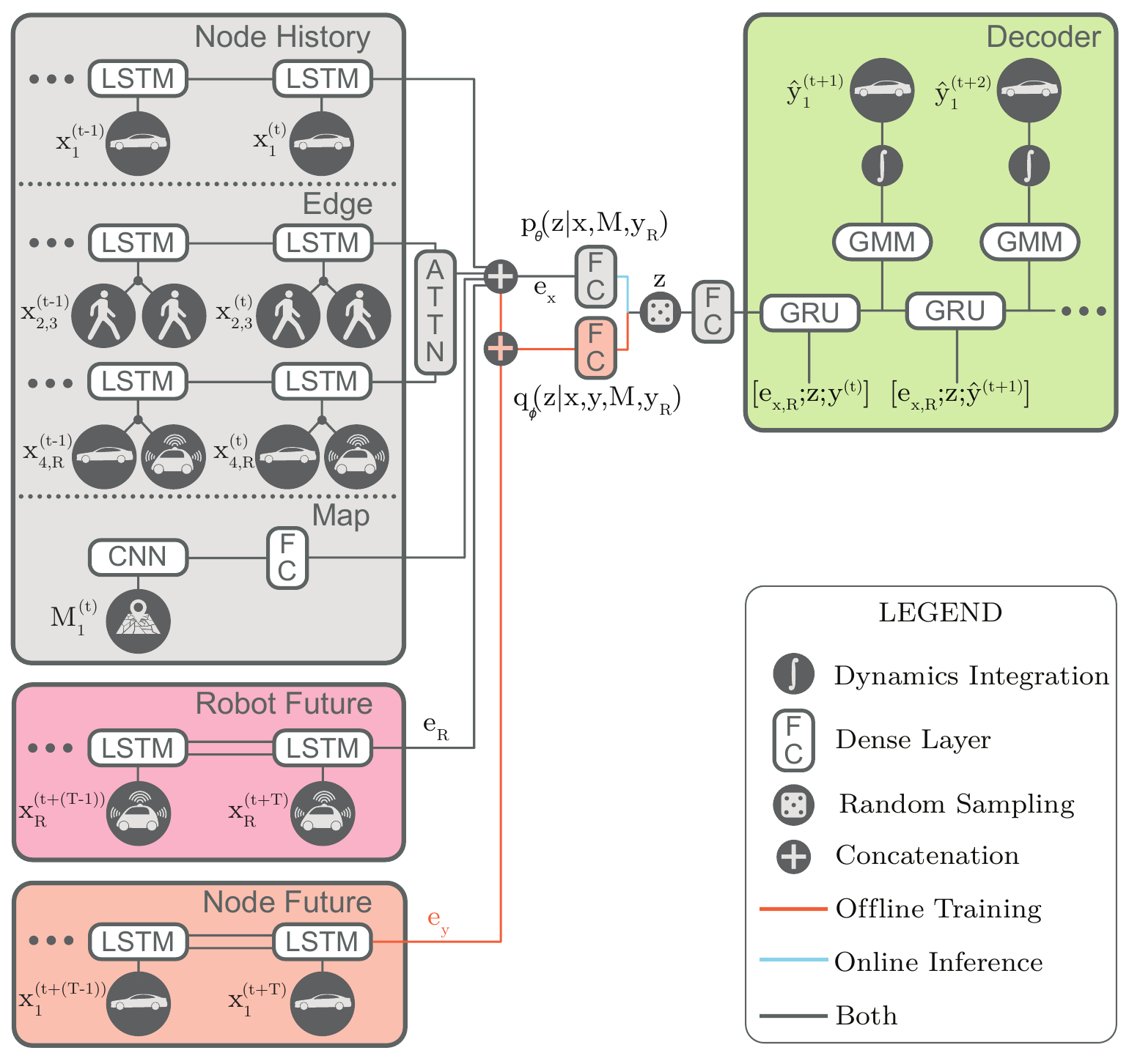}}
    \caption{\textbf{Left}: Our approach represents a scene as a directed spatiotemporal graph. Nodes and edges represent agents and their interactions, respectively. \textbf{Right}: The corresponding network architecture for Node 1.}
    \label{fig:frame_to_graph}
    \label{fig:architecture}
\end{figure}

Our approach\footnote{All of our source code, trained models, and data can be found online at\\ \url{https://github.com/StanfordASL/Trajectron-plus-plus}.} is visualized in \cref{fig:architecture}. At a high level, a spatiotemporal graph representation of the scene in question is created from its topology. Then, a similarly-structured deep learning architecture is generated that forecasts the evolution of node attributes, producing agent trajectories.

{\bf Scene Representation.}
The current scene is abstracted as a spatiotemporal graph $G = (V, E)$. Nodes represent agents and edges represent their interactions. As a result, in the rest of the paper we will use the terms ``node'' and ``agent'' interchangeably. Each node also has a semantic class matching the class of its agent (e.g., Car, Bus, Pedestrian). An edge $(A_i, A_j)$ is present in $E$ if $A_i$ influences $A_j$. In this work, the $\ell_2$ distance is used as a proxy for whether agents are influencing each other or not. Formally, an edge is directed from $A_i$ to $A_j$ if $\| \mathbf{p}_i - \mathbf{p}_j \|_2 \leq d_{S_j}$ where $\mathbf{p}_i, \mathbf{p}_j \in \mathbb{R}^2$ are the 2D world positions of agents $A_i, A_j$, respectively, and $d_{S_j}$ is a distance that encodes the perception range of agents of semantic class $S_j$. While more sophisticated methods can be used to construct edges (e.g., \cite{VemulaMuellingEtAl2018}), they usually incur extra computational overhead by requiring a complete scene graph. \cref{fig:frame_to_graph} shows an example of this scene abstraction.

We specifically choose to model the scene as a directed graph, in contrast to an undirected one as in previous approaches \cite{JainZamirEtAl2016,AlahiGoelEtAl2016,GuptaJohnsonEtAl2018,VemulaMuellingEtAl2018,IvanovicSchmerlingEtAl2018,IvanovicPavone2019}, because a directed graph can represent a more general set of scenes and interaction types, e.g., asymmetric influence. This provides the additional benefit of being able to simultaneously model agents with different perception ranges, e.g., the driver of a car looks much farther ahead on the road than a pedestrian does while walking on the sidewalk.

{\bf Modeling Agent History.} Once a graph of the scene is constructed, the model needs to encode a node's current state, its history, and how it is influenced by its neighboring nodes. To encode the observed history of the modeled agent, their current and previous states are fed into a Long Short-Term Memory (LSTM) network \cite{HochreiterSchmidhuber1997} with 32 hidden dimensions. Since we are interested in modeling trajectories, the inputs $\mathbf{x} = \mathbf{s}_{1,\dots,N(t)}^{(t - H : t)} \in \mathbb{R}^{(H+1) \times N(t) \times D}$ are the current and previous $D$-dimensional states of the modeled agents. These are typically positions and velocities, which can be easily estimated online.

Ideally, agent models should be chosen to best match their semantic class $S_i$. For example, one would usually model vehicles on the road using a bicycle model \cite{KongPfeiferEtAl2015,PadenCapEtAl2016}. However, estimating the bicycle model parameters of another vehicle from online observations is very difficult as it requires estimation of the vehicle's center of mass, wheelbase, and front wheel steer angle. As a result, in this work pedestrians are modeled as single integrators and wheeled vehicles are modeled as dynamically-extended unicycles~\cite{LaValle2006BetterUnicycle}, enabling us to account for key non-holonomic constraints (e.g., no side-slip constraints)~\cite{PadenCapEtAl2016} without requiring complex online parameter estimation procedures -- we will show through experiments that such a simplified model is already quite impactful on improving prediction accuracy.
While the dynamically-extended unicycle model serves as an important representative example, we note that our approach can also be generalized to other dynamics models, provided its parameters can either be assumed or quickly estimated online.

{\bf Encoding Agent Interactions.} To model neighboring agents' influence on the modeled agent, \emphalgname{} encodes graph edges in two steps. First, edge information is aggregated from neighboring agents of the same semantic class. In this work, an element-wise sum is used as the aggregation operation. We choose to combine features in this way rather than with concatenation or an average to handle a variable number of neighboring nodes with a fixed architecture while preserving count information \cite{BattagliaPascanuEtAl2016,IvanovicSchmerlingEtAl2018,JainZamirEtAl2016}. These aggregated states are then fed into an LSTM with 8 hidden dimensions whose weights are shared across all edge instances of the same type, e.g., all Pedestrian-Bus edge LSTMs share the same weights. Then, the encodings from all edge types that connect to the modeled node are aggregated to obtain one ``influence" representation vector, representing the effect that all neighboring nodes have. For this, an additive attention module is used~\cite{BahdanauChoEtAl2015}. Finally, the node history and edge influence encodings are concatenated to produce a single node representation vector, $e_\mathbf{x}$. 

{\bf Incorporating Heterogeneous Data.} Modern sensor suites are able to produce much more information than just tracked trajectories of other agents. Notably, HD maps are used by many real-world systems to aid localization as well as inform navigation. Depending on sensor availability and sophistication, maps can range in fidelity from simple binary obstacle maps, i.e., $M \in \{0, 1\}^{H \times W \times 1}$, to HD semantic maps, e.g., $M \in \{0, 1\}^{H \times W \times L}$ where each layer $1 \leq \ell \leq L$ corresponds to an area with semantic type (e.g., ``driveable area," ``road block," ``walkway," ``pedestrian crossing"). To make use of this information, for each modeled agent, \emphalgname{} encodes a local map, rotated to match the agent's heading, with a Convolutional Neural Network (CNN).
The CNN has 4 layers, with filters $\{5, 5, 5, 3\}$ and respective strides of $\{2, 2, 1, 1\}$. These are followed by a dense layer with 32 hidden dimensions, the output of which is concatenated with the node history and edge influence representation vectors.

More generally, one can include further additional information (e.g., raw LIDAR data, camera images, pedestrian skeleton or gaze direction estimates) in this framework by encoding it as a vector and adding it to this backbone of representation vectors, $e_\mathbf{x}$.

{\bf Encoding Future Ego-Agent Motion Plans.} Producing predictions which take into account future ego-agent motion is an important capability for robotic decision making and control. Specifically, it allows for the evaluation of a set of motion primitives with respect to possible responses from other agents. \textit{Trajectron++} can encode the future $T$ timesteps of the ego-agent's motion plan $\mathbf{y}_\text{R}$ using a bi-directional LSTM with 32 hidden dimensions. A bi-directional LSTM is used due to its strong performance on other sequence summarization tasks \cite{BritzGoldieEtAl2017}. The final hidden states are then concatenated into the backbone of representation vectors, $e_\mathbf{x}$.

{\bf Explicitly Accounting for Multimodality.} \emphalgname{} explicitly handles multimodality by leveraging the CVAE latent variable framework \cite{SohnLeeEtAl2015}. It produces the target $p(\mathbf{y} \mid \mathbf{x})$ distribution by introducing a discrete Categorical latent variable $z \in Z$ which encodes high-level latent behavior and allows for $p(\mathbf{y} \mid \mathbf{x})$ to be expressed as
    $p(\mathbf{y} \mid \mathbf{x}) = \sum_{z \in Z} p_\psi(\mathbf{y} \mid \mathbf{x}, z) p_\theta(z \mid \mathbf{x})$,
where $|Z| = 25$ and $\psi, \theta$ are deep neural network weights that parameterize their respective distributions. $z$ being discrete also aids in interpretability, as one can visualize which high-level behaviors belong to each $z$ by sampling trajectories.

During training, a bi-directional LSTM with 32 hidden dimensions is used to encode a node’s ground truth future trajectory, producing $q_\phi(z \mid \mathbf{x}, \mathbf{y})$ \cite{SohnLeeEtAl2015}.

{\bf Producing Dynamically-Feasible Trajectories.} After obtaining a latent variable $z$, it and the backbone representation vector $e_\mathbf{x}$ are fed into the decoder, a 128-dimensional Gated Recurrent Unit (GRU) \cite{ChoMerrienboerEtAl2014}. Each GRU cell outputs the parameters of a bivariate Gaussian distribution over control actions $\mathbf{u}^{(t)}$ (e.g., acceleration and steering rate).
The agent's system dynamics are then integrated with the produced control actions $\mathbf{u}^{(t)}$ to obtain trajectories in position space~\cite{Kalman1960,ThrunBurgardEtAl2005EKF}.
The only uncertainty at prediction time stems from \emphalgname{}'s output. Thus, in the case of linear dynamics (e.g., single integrators, used in this work to model pedestrians), the system dynamics are linear Gaussian. Explicitly, for a single integrator with control actions $\mathbf{u}^{(t)} = \dot{\mathbf{p}}^{(t)}$, the position mean at $t+1$ is $\mathbb{\mu}_{\mathbf{p}}^{(t+1)} = \mu_{\mathbf{p}}^{(t)} + \mu_{\mathbf{u}}^{(t)} \Delta t$, where $\mu_{\mathbf{u}}^{(t)}$ is produced by \emphalgname{}.
In the case of nonlinear dynamics (e.g., unicycle models, used in this work to model vehicles), one can still (approximately) use this uncertainty propagation scheme by linearizing the dynamics about the agent's current state and control.
Full mean and covariance equations for the single integrator and dynamically-extended unicycle models are in the appendix. In contrast to existing methods which directly output positions, our approach is uniquely able to guarantee that its trajectory samples are dynamically feasible by integrating an agent's dynamics with the predicted controls.

{\bf Output Configurations.} Based on the desired use case, \emphalgname{} can produce many different outputs. The main four are outlined below.

    1. \emph{Most Likely (ML)}: 
    The model's deterministic and most-likely single output.
    The high-level latent behavior mode and output trajectory are the modes of their respective distributions, where
    \begin{equation}
    \begin{aligned}
    z_\text{mode} &= \arg \max_z p_\theta(z \mid \mathbf{x}), \hspace{0.5cm} \mathbf{y} = \arg \max_{\mathbf{y}} p_\psi(\mathbf{y} \mid \mathbf{x}, z_\text{mode}).
    \end{aligned}
    \end{equation}
    
    2. $z_{\text{mode}}$: Predictions from the model's most-likely high-level latent behavior mode, where 
    \begin{equation}
    \begin{aligned}
    z_\text{mode} &= \arg \max_z p_\theta(z \mid \mathbf{x}), \hspace{0.5cm} \mathbf{y} \sim p_\psi(\mathbf{y} \mid \mathbf{x}, z_\text{mode}).
    \end{aligned}
    \end{equation}
    
    3. \emph{Full}: The model's full sampled output, where $z$ and $y$ are sampled sequentially according to 
    \begin{equation}
    \begin{aligned}
    z \sim p_\theta(z \mid \mathbf{x}), \hspace{0.5cm} \mathbf{y} \sim p_\psi(\mathbf{y} \mid \mathbf{x}, z).
    \end{aligned}
    \end{equation}
    
    4. \emph{Distribution}: Due to the use of a discrete latent variable and Gaussian output structure, the model can provide an analytic output distribution by directly computing $p(\mathbf{y} \mid \mathbf{x}) = \sum_{z \in Z} p_\psi(\mathbf{y} \mid \mathbf{x}, z) p_\theta(z \mid \mathbf{x})$.%

{\bf Training the Model.} We adopt the InfoVAE \cite{ZhaoSongEtAl2019} objective function, and modify it to use discrete latent states in a conditional formulation (since the model uses a CVAE). Formally, we aim to solve
\begin{equation}\label{eqn:loss_fn}
\begin{aligned}
\max_{\phi, \theta, \psi} \sum_{i=1}^N\ \mathbb{E}&_{z \sim q_\phi(\cdot \mid \mathbf{x}_i, \mathbf{y}_i)} \big[\log p_\psi(\mathbf{y}_i \mid \mathbf{x}_i, z)\big]\\
&- \beta D_{KL}\big(q_\phi(z \mid \mathbf{x}_i, \mathbf{y}_i) \parallel p_\theta(z \mid \mathbf{x}_i)\big) + \alpha I_{q} (\mathbf{x}; z),
\end{aligned}
\end{equation}
where $I_q$ is the mutual information between $\mathbf{x}$ and $z$ under the distribution $q_\phi(\mathbf{x},z)$. To compute $I_q$, we follow \cite{ZhaoSongEtAl2019} and approximate $q_\phi(z \mid \mathbf{x}_i, \mathbf{y}_i)$ with \mbox{$p_\theta(z \mid \mathbf{x}_i)$}, obtaining the unconditioned latent distribution by summing out $\mathbf{x}_i$ over the batch.
Notably, the Gumbel-Softmax reparameterization \cite{JangGuEtAl2017} is not used to backpropagate through the Categorical latent variable $z$ because it is not sampled during training time. Instead, the first term of \cref{eqn:loss_fn} is directly computed since the latent space has only $|Z| = 25$ discrete elements. Additional training details can be found in the appendix.

\section{Experiments}

Our method is evaluated on three publicly-available datasets: The ETH \cite{PellegriniEssEtAl2009}, UCY \cite{LernerChrysanthouEtAl2007}, and nuScenes \cite{CaesarBankitiEtAl2019} datasets. The ETH and UCY datasets consist of real pedestrian trajectories with rich multi-human interaction scenarios captured at 2.5 Hz ($\Delta t = 0.4s$). In total, there are 5 sets of data, 4 unique scenes, and 1536 unique pedestrians. They are a standard benchmark in the field, containing challenging behaviors such as couples walking together, groups crossing each other, and groups forming and dispersing. However, they only contain pedestrians,
so we also evaluate on the recently-released nuScenes dataset. It is a large-scale dataset for autonomous driving with 1000 scenes in Boston and Singapore. Each scene is annotated at 2 Hz ($\Delta t = 0.5s$) and is 20s long, containing up to 23 semantic object classes as well as HD semantic maps with 11 annotated layers. 
\emphalgname{} was implemented in PyTorch \cite{PaszkeGrossEtAl2017}
on a desktop computer running Ubuntu 18.04 containing an AMD Ryzen 1800X CPU and two NVIDIA GTX 1080 Ti GPUs.
We trained the model for 100 epochs ($\sim 3$ hours) on the pedestrian datasets and 12 epochs ($\sim 8$ hours) on the nuScenes dataset.

{\bf Evaluation Metrics.} As in prior work \cite{AlahiGoelEtAl2016,GuptaJohnsonEtAl2018,IvanovicPavone2019,SadeghianKosarajuEtAl2019,KosarajuSadeghianEtAl2019,ZhaoXuEtAl2019}, our method for trajectory forecasting is evaluated with the following four error metrics:

    1. \textit{Average Displacement Error (ADE)}: Mean $\ell_2$ distance between the ground truth and predicted trajectories.
    
    2. \textit{Final Displacement Error (FDE)}: $\ell_2$ distance between the predicted final position and the ground truth final position at the prediction horizon $T$.
    
    3. \textit{Kernel Density Estimate-based Negative Log Likelihood (KDE NLL)}: Mean NLL of the ground truth trajectory under a distribution created by fitting a kernel density estimate on trajectory samples~\cite{IvanovicPavone2019,ThiedeBrahma2019}.
    
    4. \textit{Best-of-N (BoN)}: The minimum ADE and FDE from $N$ randomly-sampled trajectories.
We compare our method to an exhaustive set of state-of-the art deterministic and generative approaches.

{\bf Deterministic Baselines.} Our method is compared against the following deterministic baselines: (1) \textit{Linear}: A linear regressor with parameters estimated by minimizing least square error. (2) \textit{LSTM}: An LSTM network with only agent history information. (3) \textit{Social LSTM} \cite{AlahiGoelEtAl2016}: Each agent is modeled with an LSTM and nearby agents' hidden states are pooled at each timestep using a proposed social pooling operation. (4) \textit{Social Attention} \cite{VemulaMuellingEtAl2018}: Same as \cite{AlahiGoelEtAl2016}, but all other agents' hidden states are incorporated via a proposed social attention operation.

{\bf Generative Baselines.} On the ETH and UCY datasets, our method is compared against the following generative baselines: 
(1) \textit{S-GAN} \cite{GuptaJohnsonEtAl2018}: Each agent is modeled with an LSTM-GAN, which is an LSTM encoder-decoder whose outputs are the generator of a GAN. The generated trajectories are then evaluated against the ground truth trajectories with a discriminator. 
(2) \textit{SoPhie} \cite{SadeghianKosarajuEtAl2019}: An LSTM-GAN with the addition of a proposed physical and social attention module. 
(3) \textit{MATF} \cite{ZhaoXuEtAl2019}: An LSTM-GAN model that leverages CNNs to fuse agent relationships and encode environmental information. 
(4) \textit{Trajectron}~\cite{IvanovicPavone2019}: An LSTM-CVAE encoder-decoder which is explicitly constructed to match the spatiotemporal structure of the scene. Its scene abstraction is similar to ours, but uses undirected edges.

On the nuScenes dataset, the following methods are also compared against: 
(5) \textit{Convolutional Social Pooling (CSP)} \cite{DeoTrivedi2018}: An LSTM-based approach which explicitly considers a fixed number of movement classes and predicts which of those the modeled agent is likely to take. 
(6) \textit{CAR-Net} \cite{SadeghianLegrosEtAl2018}: An LSTM-based approach which encodes scene context with visual attention. 
(7) \textit{SpAGNN} \cite{CasasGulinoEtAl2019}: A CNN encodes raw LIDAR and semantic map data to produce object detections, from which a Graph Neural Network (GNN) produces probabilistic, interaction-aware trajectories.

{\bf Evaluation Methodology.} For the ETH and UCY datasets, a leave-one-out strategy is used for evaluation, similar to previous works~\cite{AlahiGoelEtAl2016,GuptaJohnsonEtAl2018,IvanovicPavone2019,KosarajuSadeghianEtAl2019,SadeghianKosarajuEtAl2019,ZhaoXuEtAl2019}, where the model is trained on four datasets and evaluated on the held-out fifth. An observation length of 8 timesteps (3.2s) and a prediction horizon of 12 timesteps (4.8s) is used for evaluation. 
For the nuScenes dataset, we split off 15\% of the train set for hyperparameter tuning and test on the provided validation set.

Throughout the following, we report the performance of \emphalgname{} in multiple configurations. Specifically, \textit{Ours} refers to the base model using only node and edge encoding, trained to predict agent velocities and Euler integrating velocity to produce positions; 
\textit{Ours}+$\int$ is the base model with dynamics integration, trained to predict control actions and integrating the agent's dynamics with the control actions to produce positions; 
\textit{Ours}+$\int,M$ additionally includes the map encoding CNN; and \textit{Ours}+$\int,M,\mathbf{y}_\text{R}$ adds the robot future encoder.

\begin{table}[t]
\fontsize{8}{8}\selectfont
\centering
\caption{
\textbf{(a)} Our model's deterministic Most Likely output outperforms other deterministic methods on displacement error metrics, even if it was not originally trained to do so.
\textbf{(b)} Our model's probabilistic Full output significantly outperforms other methods, yielding accurate predictions even in a small number of samples.
Lower is better. Bold indicates best.}
\begin{tabular}{l|cccc|cc}
\toprule
\multicolumn{1}{c|}{\multirow{2}{*}{\textbf{Dataset}}} & \multicolumn{6}{c}{(a) \textbf{ADE/FDE (m)}} \Bstrut \\ \cline{2-7} 
\multicolumn{1}{c|}{}                         & Linear      & LSTM        & S-LSTM \cite{GuptaJohnsonEtAl2018} & S-ATTN \cite{VemulaMuellingEtAl2018} & Ours (ML) & Ours+$\int$ (ML) \TBstrut \\ \midrule
ETH                                           & $1.33$/$2.94$ & $1.09$/$2.41$ & $1.09$/$2.35$ & $\mathbf{0.39}$/$3.74$      & $0.71$/$\mathbf{1.66}$  & $0.71$/$1.68$             \\
Hotel                                         & $0.39$/$0.72$ & $0.86$/$1.91$ & $0.79$/$1.76$ & $0.29$/$2.64$      & $\mathbf{0.22}$/$\mathbf{0.46}$  & $\mathbf{0.22}$/$\mathbf{0.46}$             \\
Univ                                          & $0.82$/$1.59$ & $0.61$/$1.31$ & $0.67$/$1.40$ & $\mathbf{0.33}$/$3.92$      & $0.44$/$1.17$  & $0.41$/$\mathbf{1.07}$             \\
Zara 1                                        & $0.62$/$1.21$ & $0.41$/$0.88$ & $0.47$/$1.00$ & $\mathbf{0.20}$/$\mathbf{0.52}$      &  $0.30$/$0.79$ & $0.30$/$0.77$             \\
Zara 2                                        & $0.77$/$1.48$ & $0.52$/$1.11$ & $0.56$/$1.17$ & $0.30$/$2.13$      & $\mathbf{0.23}$/$\mathbf{0.59}$  & $\mathbf{0.23}$/$\mathbf{0.59}$             \\ \midrule
Average                                       & $0.79$/$1.59$ & $0.70$/$1.52$ & $0.72$/$1.54$ & $\mathbf{0.30}$/$2.59$      &  $0.38$/$0.93$ & $0.37$/$\mathbf{0.91}$  \Bstrut   \\ %
\hline
\hline
\multicolumn{1}{c|}{\multirow{2}{*}{\textbf{Dataset}}} & \multicolumn{6}{c}{(b) \textbf{ADE/FDE, Best of 20 Samples (m)}}  \TBstrut \\ \cline{2-7} 
\multicolumn{1}{c|}{}                         & S-GAN \cite{GuptaJohnsonEtAl2018} & SoPhie \cite{SadeghianKosarajuEtAl2019} & Trajectron \cite{IvanovicPavone2019} & MATF \cite{ZhaoXuEtAl2019} & Ours (Full) & Ours+$\int$ (Full) \TBstrut \\ \midrule
ETH                                           & $0.81$/$1.52$  & $0.70$/$1.43$   & $0.59$/$1.14$  & $1.01$/$1.75$ & $\mathbf{0.39}$/$\mathbf{0.83}$ & $0.43$/$0.86$  \\
Hotel                                         & $0.72$/$1.61$  & $0.76$/$1.67$   & $0.35$/$0.66$  & $0.43$/$0.80$ & $\mathbf{0.12}$/$0.21$ & $\mathbf{0.12}$/$\mathbf{0.19}$  \\
Univ                                          & $0.60$/$1.26$  & $0.54$/$1.24$   & $0.54$/$1.13$  & $0.44$/$0.91$ & $\mathbf{0.20}$/$0.44$ & $0.22$/$\mathbf{0.43}$ \\
Zara 1                                        & $0.34$/$0.69$  & $0.30$/$0.63$   & $0.43$/$0.83$  & $0.26$/$0.45$ & $\mathbf{0.15}$/$0.33$ & $0.17$/$\mathbf{0.32}$ \\
Zara 2                                        & $0.42$/$0.84$  & $0.38$/$0.78$   & $0.43$/$0.85$  & $0.26$/$0.57$ & $\mathbf{0.11}$/$\mathbf{0.25}$ & $0.12$/$\mathbf{0.25}$  \\ \midrule
Average                                       & $0.58$/$1.18$  & $0.54$/$1.15$   & $0.47$/$0.92$  & $0.48$/$0.90$ & $\mathbf{0.19}$/$\mathbf{0.41}$ & $0.21$/$\mathbf{0.41}$  \\ \bottomrule
\end{tabular}
\label{tab:deterministic_fde}
\label{tab:generative_BoN}

Legend: $\int$ = Integration via Dynamics, $M$ = Map Encoding, $\mathbf{y}_\text{R}$ = Robot Future Encoding
\end{table}

\subsection{ETH and UCY Datasets}

Our approach is first evaluated on the ETH \cite{PellegriniEssEtAl2009} and UCY \cite{LernerChrysanthouEtAl2007} Pedestrian Datasets, against deterministic methods on standard trajectory forecasting metrics. It is difficult to determine the current state-of-the-art in deterministic methods as there are contradictions between the results reported by the same authors in \cite{GuptaJohnsonEtAl2018} and \cite{AlahiGoelEtAl2016}. In Table 1 of \cite{AlahiGoelEtAl2016}, Social LSTM \textit{convincingly} outperforms a baseline LSTM without pooling. However, in Table 1 of \cite{GuptaJohnsonEtAl2018}, Social LSTM is actually \textit{worse} than the same baseline on average. Thus, when comparing against Social LSTM we report the results summarized in Table 1 of \cite{GuptaJohnsonEtAl2018} as it is the most recent work by the same authors. Further, the values reported by Social Attention in \cite{VemulaMuellingEtAl2018} seem to have unusually high ratios of FDE to ADE.
Nearly every other method (including ours) has FDE/ADE ratios around $2-3\times$ whereas Social Attention's are around $3-12\times$. Social Attention's errors on the Univ dataset are especially striking, as its FDE of $3.92$ is $12\times$ its ADE of $0.33$, meaning its prediction error on the other 11 timesteps is essentially zero. We still compare against the values reported in \cite{VemulaMuellingEtAl2018} as there is no publicly-released code, but this raises doubts of their validity. To fairly compare against prior work, neither map encoding nor future motion plan encoding is used.
Only the node history and edge encoders are used in the model's encoder. Additionally, the model's deterministic ML output scheme is employed, which produces the model's most likely single trajectory. \cref{tab:deterministic_fde}~(a) summarizes these results and shows that our approach is competitive with state-of-the-art deterministic regressors on displacement error metrics (outperforming existing approaches by $33\%$ on mean FDE), even though our method was not originally trained to minimize this.
It makes sense that the model performs similarly with and without dynamics integration for pedestrians, since they are modeled as single integrators. Thus, their control actions are velocities which matches the base model's output structure.

\begin{table}[t]
\fontsize{8}{8}\selectfont
\centering
\caption{Mean KDE-based NLL for each dataset. Lower is better. 2000 trajectories were sampled per model at each prediction timestep. Bold indicates the best values.}
\begin{tabular}{l|cc|cc}
\toprule
\multicolumn{1}{c|}{\multirow{2}{*}{\textbf{Dataset}}} & \multicolumn{4}{c}{\textbf{KDE NLL}} \\ \cline{2-5} 
\multicolumn{1}{c|}{}                         & S-GAN \cite{GuptaJohnsonEtAl2018}       & Trajectron \cite{IvanovicPavone2019}     & Ours (Full)  & Ours+$\int$ (Full) \Tstrut  \\ \midrule
ETH                                           & $15.70$       & $2.99$            & $1.80$ & $\mathbf{1.31}$      \\ 
Hotel                                         & $8.10$        & $2.26$            & $-1.29$ & $\mathbf{-1.94}$     \\ 
Univ                                          & $2.88$        & $1.05$            & $-0.89$ & $\mathbf{-1.13}$      \\ 
Zara 1                                        & $1.36$        & $1.86$            & $-1.13$  & $\mathbf{-1.41}$      \\ 
Zara 2                                        & $0.96$        & $0.81$            & $-2.19$ & $\mathbf{-2.53}$      \\ \midrule
Average                                       & $5.80$        & $1.79$            & $-0.74$ & $\mathbf{-1.14}$      \\ \bottomrule
\end{tabular}
\label{tab:kde_nll}

Legend: $\int$ = Integration via Dynamics, $M$ = Map Encoding, $\mathbf{y}_\text{R}$ = Robot Future Encoding
\end{table}

To more concretely compare generative methods, we use the KDE-based NLL metric proposed in \cite{IvanovicPavone2019,ThiedeBrahma2019}, an approach that maintains full output distributions and compares the log-likelihood of the ground truth under different methods' outputs. \cref{tab:kde_nll} summarizes these results and shows that our method significantly outperforms others. This is also where the performance improvements brought by the dynamics integration scheme are clear. It yields the best performance because the model is now explicitly trained on the distribution it is seeking to output (the loss function term $p_\psi(\mathbf{y} | \mathbf{x}, z)$ is now directly over positions), whereas the base model is trained on velocity distributions, the integration of which (with no accounting for system dynamics) introduces errors.
Unfortunately, at this time there are no publicly-released models for SoPhie~\cite{SadeghianKosarajuEtAl2019} or MATF~\cite{ZhaoXuEtAl2019}, so they cannot be evaluated with the KDE-based NLL metric. Instead, we evaluate \emphalgname{} with the Best-of-$N$ metric used in their works.
\cref{tab:generative_BoN}~(b) summarizes these results, and shows that our method \textit{significantly} ourperforms the state-of-the-art \cite{ZhaoXuEtAl2019}, achieving $55-60\%$ lower average errors.

{\bf Map Encoding.} To evaluate the effect of incorporating heterogeneous data, we compare the performance of \emphalgname{} with and without the map encoder. Specifically, we compare the frequency of obstacle violations in 2000 trajectory samples from the Full model output on the ETH - University scene, which provides a simple binary obstacle map. Overall, our approach generates colliding predictions $1.0\%$ of the time with map encoding, compared to $4.6\%$ without map encoding.
We also study how much of a reduction there is for pedestrians that are especially close to an obstacle (i.e. they have at least one obstacle-violating trajectory in their Full output), an example of which is shown in the appendix.
In this regime, our approach generates colliding predictions $4.9\%$ of the time with map encoding, compared to $21.5\%$ without map encoding.

\subsection{nuScenes Dataset}

\begin{table}[t]
\fontsize{6.5}{6.5}\selectfont
\centering
\caption{\textbf{[nuScenes]} \textbf{(a)}: Vehicle-only FDE across time for \emphalgname{}
compared to that of other single-trajectory and probabilistic approaches. Bold indicates best. \textbf{(b)}:~Pedestrian-only FDE and KDE NLL across time for \emphalgname{}.
}
\begin{tabular}{l|cccc}
\toprule
\multicolumn{5}{c}{(a) \textbf{Vehicle-only}}\\
\midrule
\multicolumn{1}{c|}{\multirow{2}{*}{\textbf{Method}}} & \multicolumn{4}{c}{\textbf{FDE (m)}} \Bstrut \\
\cline{2-5} & @1s & @2s & @3s & @4s \Tstrut \\
\midrule
Const. Velocity & $0.32$ & $0.89$ & $1.70$ & $2.73$\\
S-LSTM$^*$~\cite{AlahiGoelEtAl2016,CasasGulinoEtAl2019} & $0.47$ & - & $1.61$ & - \\
CSP$^*$~\cite{DeoTrivedi2018,CasasGulinoEtAl2019} & $0.46$ & - & $1.50$ & - \\
CAR-Net$^*$~\cite{SadeghianLegrosEtAl2018,CasasGulinoEtAl2019} & $0.38$ & - & $1.35$ & - \\
SpAGNN$^*$~\cite{CasasGulinoEtAl2019} & $0.36$ & - & $1.23$ & - \\
\midrule
Ours (ML) & $0.18$ & $0.57$ & $1.25$ & $2.24$\\
Ours+$\int$,$M$ (ML) & $\mathbf{0.07}$ & $\mathbf{0.45}$ & $\mathbf{1.14}$ & $\mathbf{2.20}$\\
\bottomrule
\end{tabular}
\hfill%
\begin{tabular}{l|cccc|cccc}
\toprule
\multicolumn{9}{c}{(b) \textbf{Pedestrian-only}}\\
\midrule
\multicolumn{1}{c|}{\multirow{2}{*}{\textbf{Method}}} & \multicolumn{4}{c|}{\textbf{KDE NLL}} & \multicolumn{4}{c}{\textbf{FDE (m)}} \Bstrut \\
\cline{2-9} & @1s & @2s & @3s & @4s & @1s & @2s & @3s & @4s \Tstrut \\
\midrule
Ours (ML) & $-2.69$ & $-2.46$ & $-1.76$ & $-1.09$ & $0.03$ & $0.17$ & $0.37$ & $0.60$\\
Ours+$\int$,$M$ (ML) & $-5.58$  & $-3.96$ & $-2.77$  & $-1.89$ & $0.01$ & $0.17$ & $0.37$ & $0.62$ \\
\bottomrule
\end{tabular}
\label{tab:nuscenes_fde}
\label{tab:nuscenes_ped}

\fontsize{7}{7}\selectfont
$^*$We subtracted $22$-$24cm$ from these reported values (their detection/tracking error~\cite{CasasGulinoEtAl2019}), as we do not use a detector/tracker. This is done to establish a fair comparison.

Legend: $\int$ = Integration via Dynamics, $M$ = Map Encoding, $\mathbf{y}_\text{R}$ = Robot Future Encoding.
\end{table}

To further evaluate the model's ability to use heterogeneous data and simultaneously model multiple semantic classes of agents, we evaluate it on the nuScenes dataset~\cite{CaesarBankitiEtAl2019}. Again, the deterministic ML output scheme is used to fairly compare with other single-trajectory predictors.
The trajectories of both Pedestrians and Cars are forecasted, two semantic object classes which 
account for most of the 23 possible object classes present in the dataset. To obtain an estimate of prediction quality degradation over time, we compute the model's FDE at $t = \{1, 2, 3, 4\}s$ for all tracked objects with at least $4s$ of available future data. We also implement a constant velocity baseline, which simply maintains the agent's heading and speed for the prediction horizon. \cref{tab:nuscenes_fde} (a) summarizes the model's performance in comparison with state-of-the-art vehicle trajectory prediction models. Since other methods use a detection/tracking module (whereas ours does not), to establish a fair comparison we subtracted other methods' detection and tracking error from their reported values.
The dynamics integration scheme and map encoding yield a noticeable improvement with vehicles, as their dynamically-extended unicycle dynamics now differ from the single integrator assumption made by the base model.
Note that our method was only trained to predict $3s$ into the future, thus its performance at $4s$ also provides a measure of its capability to generalize beyond its training configuration. Other methods do not report values at $2$s and $4$s. As can be seen, \emphalgname{} outperforms existing approaches without facing a sharp degradation in performance after $3s$.
Our approach's performance on pedestrians is reported in \cref{tab:nuscenes_ped} (b), where the inclusion of HD maps and dynamics integration similarly improve performance as in the pedestrian datasets.

\begin{table}[t]
\fontsize{8}{8}\selectfont
\centering
\caption{\textbf{[nuScenes]} 
\textbf{(a)}: Vehicle-only prediction performance
for ablated versions of our model. \textbf{(b)}: The same, but excluding the ego-robot from consideration (as it is being conditioned on). This shows that our model's robot future conditional performance does not arise from merely removing the ego-vehicle.
}
\begin{tabular}{ccc|rrrr|cccc|cccc}
\toprule
 \multicolumn{15}{c}{(a) \textbf{Including the Ego-Vehicle}} \\
 \midrule
\multicolumn{3}{c|}{\textbf{Ablation}}   & \multicolumn{4}{c|}{\textbf{KDE NLL}}  & \multicolumn{4}{c|}{\textbf{FDE ML (m)}}  & \multicolumn{4}{c}{\textbf{B. Viol. (\%)}} \\
$\int$      & $M$         & $\mathbf{y}_\text{R}$ & @1s & @2s  & @3s  & @4s   & @1s   & @2s  & @3s   & @4s   & @1s & @2s   & @3s  & @4s \\ \midrule
    -       &      -      &     -          &  $0.81$   &  $0.05$  &   $0.37$   &   $0.87$    &   $0.18$    &   $0.57$   &   $1.25$    &   $2.24$    &  $0.2$  &   $0.6$   &  $2.8$   &   $6.9$ \\
 \checkmark &      -      &     -          &  $-4.28$   &   $-2.82$    &   $-1.67$   &  $-0.76$    &   $0.07$    &   $0.45$   &   $1.13$    &   $2.17$    &  $0.2$   &   $0.7$    &   $3.2$   &  $8.1$   \\
 \checkmark & \checkmark  &     -          &   $-4.17$  &   $-2.74$   &   $-1.62$   &    $-0.70$   &    $0.07$   &   $0.45$   &    $1.14$   &   $2.20$    &   $0.3$  &    $0.6$   &   $2.8$   &  $7.6$   \\
 \hline
 \hline 
 \multicolumn{15}{c}{(b) \textbf{Excluding the Ego-Vehicle}} \Tstrut \\
 \midrule
\multicolumn{3}{c|}{\textbf{Ablation}}   & \multicolumn{4}{c|}{\textbf{KDE NLL}}  & \multicolumn{4}{c|}{\textbf{FDE ML (m)}}  & \multicolumn{4}{c}{\textbf{B. Viol. (\%)}} \\
$\int$      & $M$         & $\mathbf{y}_\text{R}$ & @1s & @2s  & @3s  & @4s   & @1s  & @2s  & @3s & @4s & @1s & @2s  & @3s  & @4s \\ \midrule
\checkmark  & \checkmark  &   -            &  $-4.26$   &   $-2.86$   &  $-1.76$    &   $-0.87$    &   $0.07$   &   $0.44$   &  $1.09$   &  $2.09$   &  $0.3$   &   $0.6$   &  $2.8$    &  $7.6$   \\
\checkmark  & \checkmark  &   \checkmark   &   $-3.90$  &   $-2.76$   &  $-1.75$    &   $-0.93$    &   $0.08$    &   $0.34$   &   $0.81$    &    $1.50$   &  $0.3$   &   $0.5$    &   $1.6$   &  $4.2$   \\
\bottomrule
\end{tabular}\label{tab:nuscenes_abl}
Legend: $\int$ = Integration via Dynamics, $M$ = Map Encoding, $\mathbf{y}_\text{R}$ = Robot Future Encoding
\end{table}

{\bf Ablation Study.} To develop an understanding of which model components influence performance, a comprehensive ablation study is performed in \cref{tab:nuscenes_abl}.
As can be seen in the first row,
even the base model's deterministic ML output performs strongly relative to current state-of-the-art approaches for vehicle trajectory forecasting~\cite{CasasGulinoEtAl2019}. 
Adding the dynamics integration scheme yields a drastic reduction in NLL as well as FDE at all prediction horizons. There is also an associated slight increase in the frequency of road boundary-violating predictions. This is a consequence of training in position (as opposed to velocity) space, which yields more variability in the corresponding predictions.
Additionally including map encoding maintains prediction accuracy while reducing the frequency of boundary-violating predictions.
The effect of conditioning on the ego-vehicle's future motion plan is also studied, with results summarized in \cref{tab:nuscenes_abl} (b). 
As one would expect, providing the model with future motion plans of the ego-vehicle yields significant reductions in error and road boundary violations. 
This use-case is common throughout autonomous driving as the ego-vehicle repeatedly produces future motion plans at every timestep by evaluating motion primitives. Overall, dynamics integration is the dominant performance-improving module.%

\begin{figure}[t]
    \centering
    \raisebox{-0.5\height}{\begin{overpic}[width=0.29\linewidth,frame]{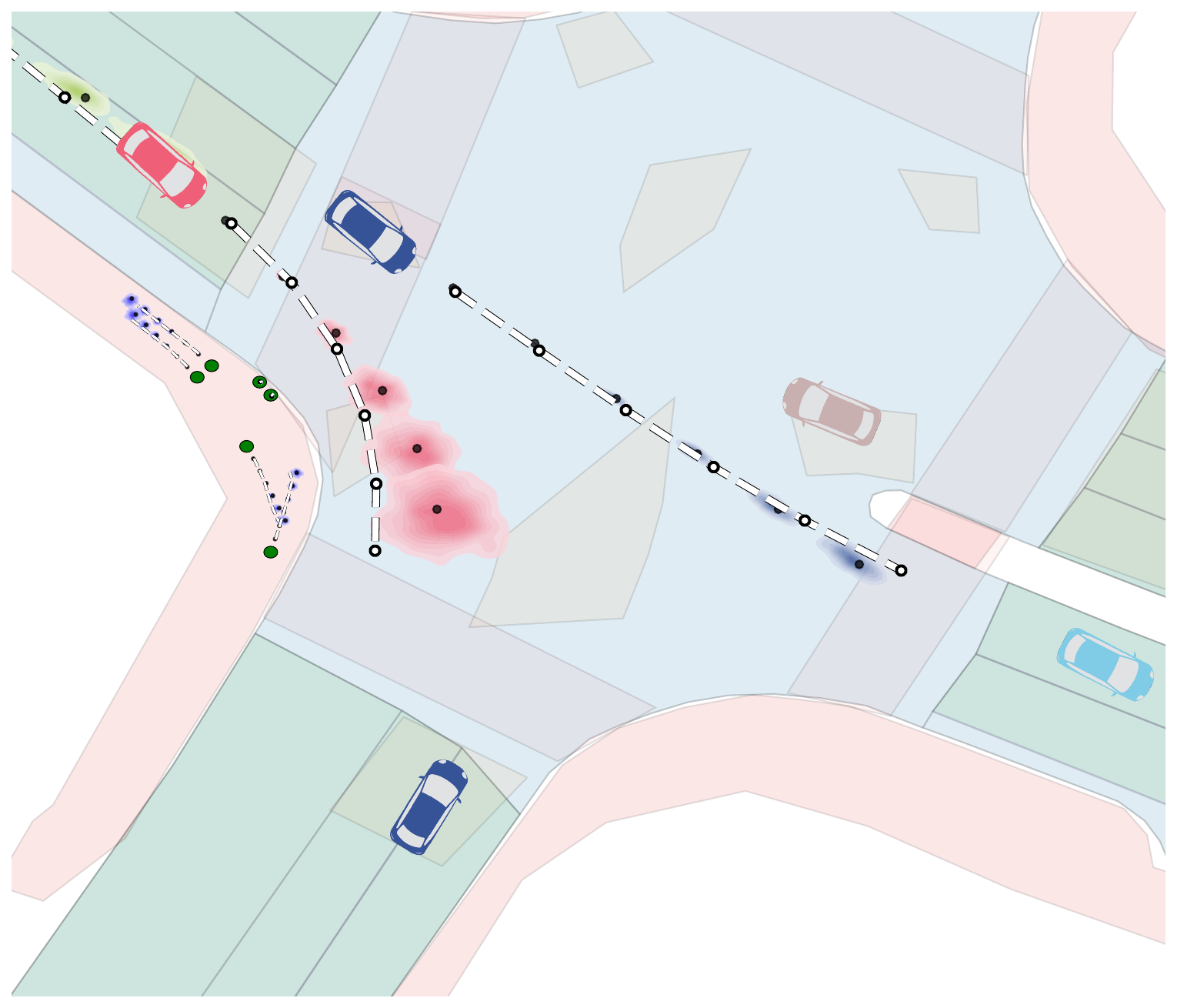}%
    \put (45, 5) {(a) Ours}%
    \end{overpic}}
    \raisebox{-0.5\height}{\begin{overpic}[width=0.29\linewidth,frame]{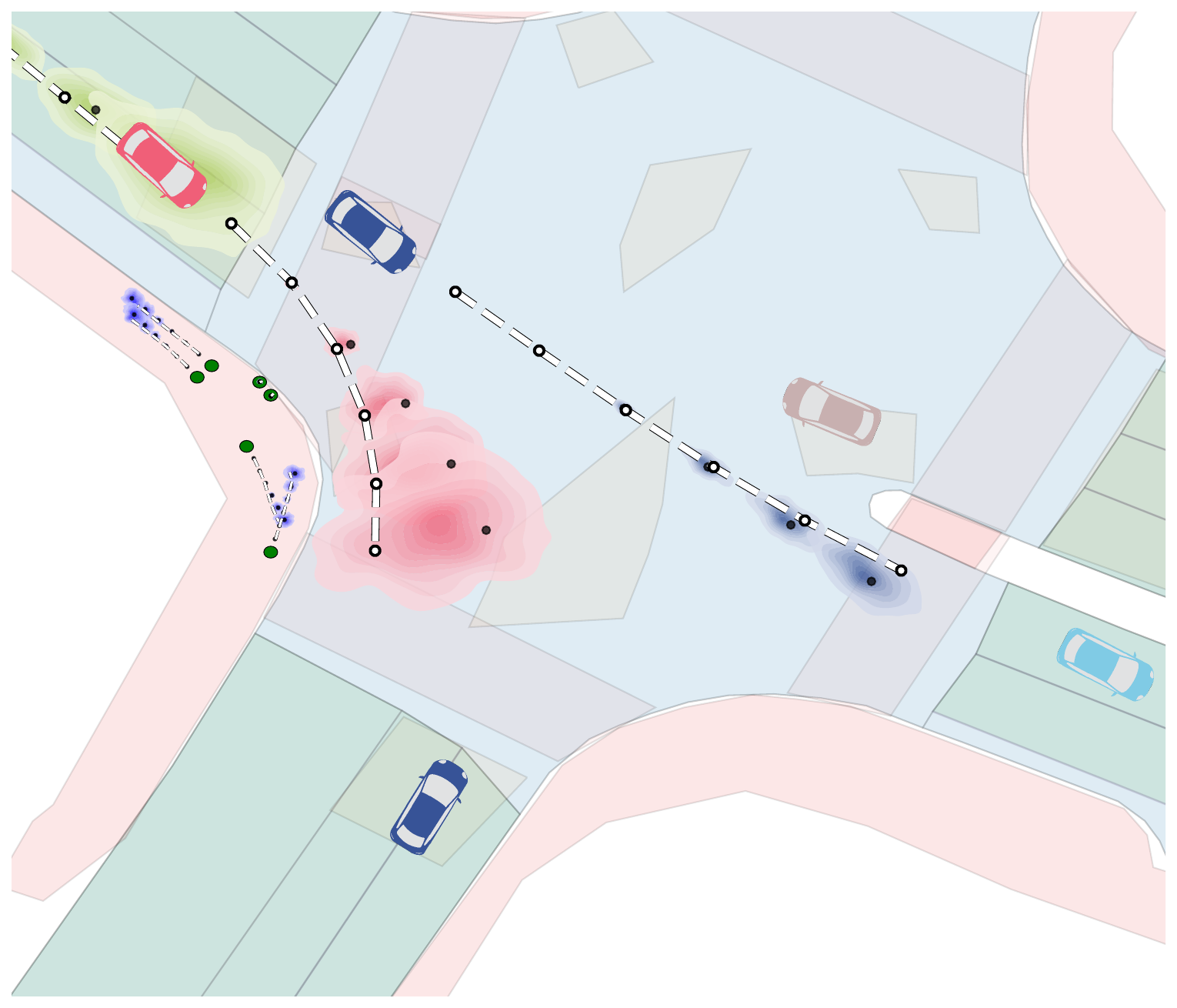}%
    \put (45, 5) {(b)+$\int$}%
    \end{overpic}}
    \raisebox{-0.5\height}{\begin{overpic}[width=0.29\linewidth,,frame]{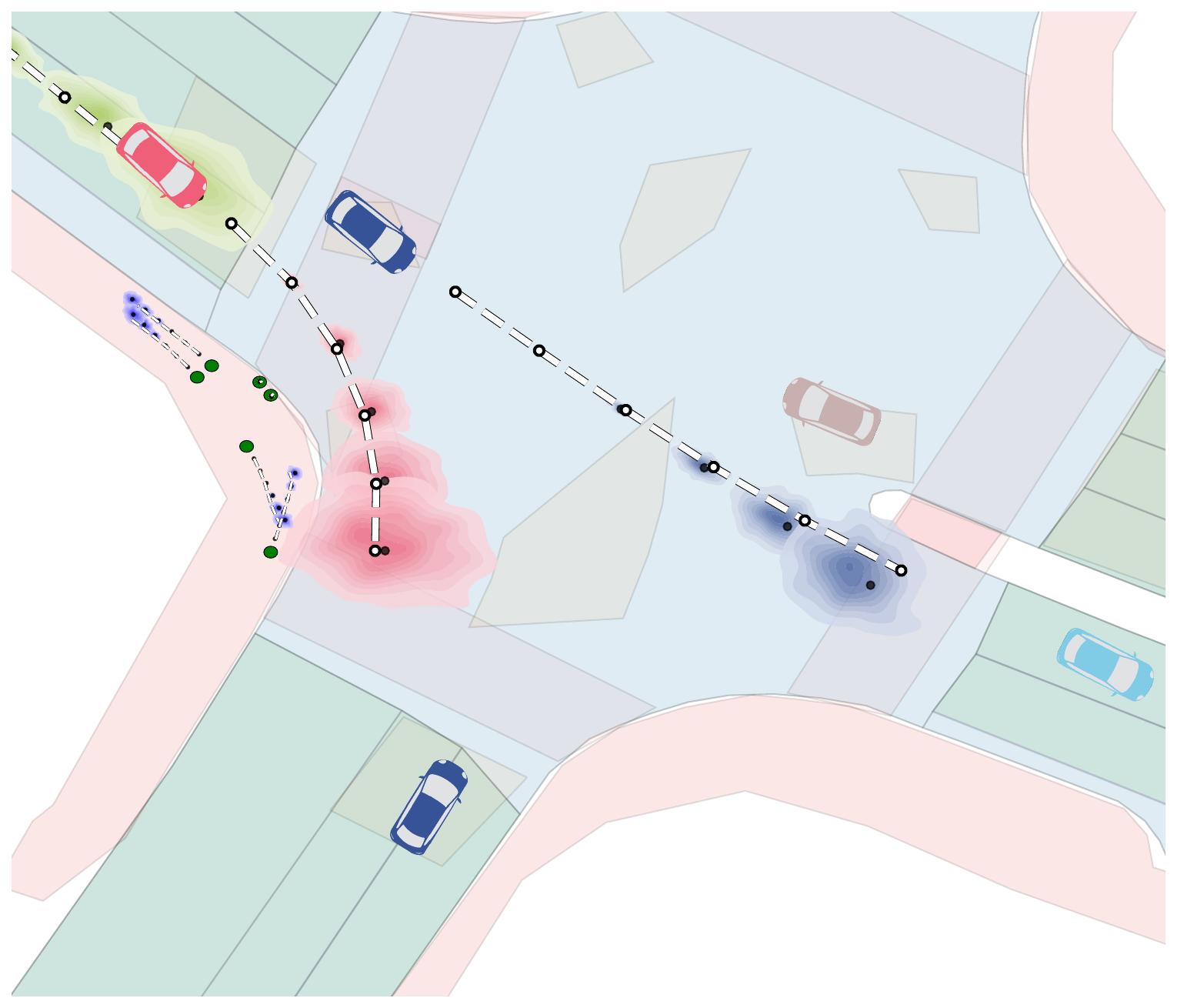}%
    \put (45, 5) {(c)+$\int, M$}%
    \end{overpic}}
    \raisebox{-0.5\height}{\begin{overpic}[width=0.1\linewidth,]{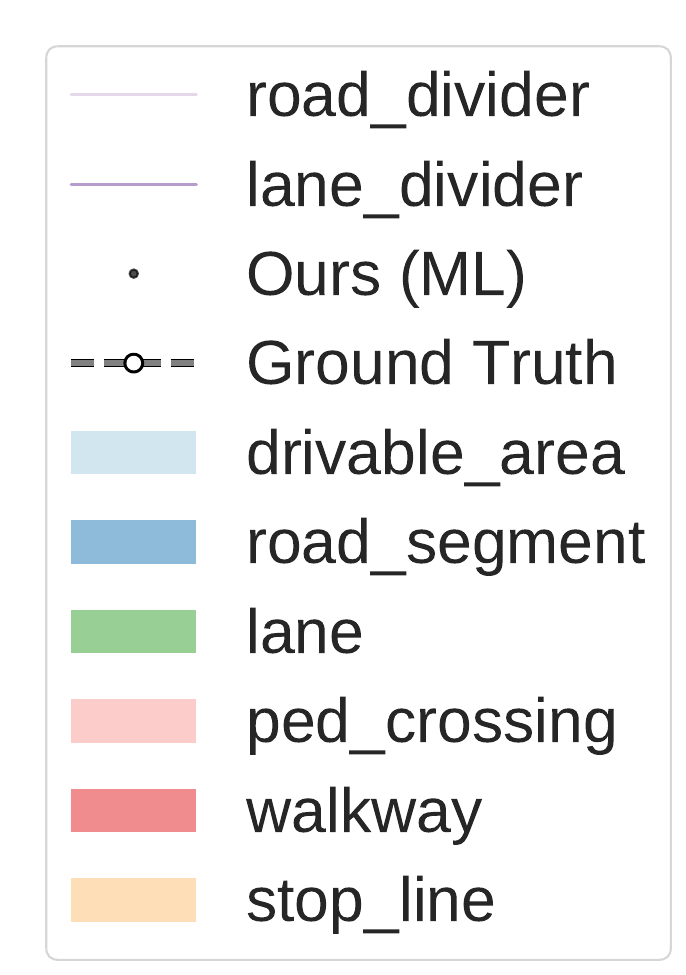}%
    \end{overpic}}
    \caption{\textbf{[nuScenes]} The same scene as forecast by three versions of \emphalgname{}. \textbf{(a)}~The base model tends to under-shoot turns, and makes overly-confident predictions. 
    \textbf{(b)}~Our approach better captures position uncertainty with dynamics integration, producing well-calibrated probabilities.
    \textbf{(c)}~The model is able to leverage the additional information that a map provides, yielding accurate predictions.}
    \label{fig:nuscenes_map_quali}
    \label{fig:nuscenes_robot_quali}
\end{figure}

{\bf Qualitative Comparison.} \cref{fig:nuscenes_map_quali} shows trajectory predictions from the base model, with dynamics integration, and with dynamics integration + map encoding. 
In it, one can see that the base model (predicting in velocity space) undershoots the turn for the red car, predicting that it will end up in oncoming traffic. 
With the integration of dynamics, the model captures multimodality in the agent's action, predicting both the possibility of a right turn and continuing straight.
With the addition of map encoding, the predictions are not only more accurate, but nearly all probability mass now lies within the correct side of the road. This is in contrast to versions of the model without map encoding which predict that the red car might move into oncoming traffic.

{\bf Online Runtime.} A key consideration in 
robotics is runtime complexity. As a result, we evaluate the time it takes \emphalgname{} to perform forward inference on commodity hardware. The results are summarized in the appendix, and confirm that our model scales well to scenes with many agents and interactions.

\section{Conclusion}
In this work, we present \emphalgname{}, a generative multi-agent trajectory forecasting approach which uniquely addresses our desiderata for an open, generally-applicable, and extensible framework. It can incorporate heterogeneous data beyond prior trajectory information and is able to produce future-conditional predictions that respect dynamics constraints, all
while producing full probability distributions, which are especially useful in downstream robotic tasks such as motion planning, decision making, and control. 
It achieves state-of-the-art prediction performance in a variety of metrics on standard and new real-world multi-agent human behavior datasets.

{\bf Acknowledgment.} 
Tim Salzmann is supported by a fellowship within the IFI programme of the German Academic Exchange Service (DAAD). We thank Matteo Zallio for his help in visually communicating our work and Amine Elhafsi for sharing his dynamics knowledge and proofreading. We also thank Brian Yao for improving our pedestrian dataset evaluation script, Osama Makansi for improving our result reporting, and Lu Zhang for correcting typos in our references. This work was supported in part by the Ford-Stanford Alliance. This article solely reflects the opinions and conclusions of its authors.

\nocite{HigginsMattheyEtAl2017,BowmanVilnisEtAl2015,HallacLeskovecEtAl2015,LaValle2006Unicycle,SchollerAravantinosEtAl2020}

\bibliographystyle{splncs04}
\bibliography{ASL_papers,main}

\newcommand{\noopsort}[1]{} \newcommand{\printfirst}[2]{#1}
  \newcommand{\singleletter}[1]{#1} \newcommand{\switchargs}[2]{#2#1}
\begin{thebibliography}{10}
\providecommand{\url}[1]{\texttt{#1}}
\providecommand{\urlprefix}{URL }
\providecommand{\doi}[1]{https://doi.org/#1}

\bibitem{AlahiGoelEtAl2016}
Alahi, A., Goel, K., Ramanathan, V., Robicquet, A., Fei-Fei, L., Savarese, S.:
  Social {LSTM}: Human trajectory prediction in crowded spaces. In: {IEEE
  Conf.\ on Computer Vision and Pattern Recognition} (2016)

\bibitem{BahdanauChoEtAl2015}
Bahdanau, D., Cho, K., Bengio, Y.: Neural machine translation by jointly
  learning to align and translate. In: {Int.\ Conf.\ on Learning
  Representations} (2015)

\bibitem{BattagliaPascanuEtAl2016}
Battaglia, P.W., Pascanu, R., Lai, M., Rezende, D., Kavukcuoglu, K.:
  Interaction networks for learning about objects, relations and physics. In:
  {Conf.\ on Neural Information Processing Systems} (2016)

\bibitem{BowmanVilnisEtAl2015}
Bowman, S.R., Vilnis, L., Vinyals, O., Dai, A.M., Jozefowicz, R., Bengio, S.:
  Generating sentences from a continuous space. In: {Proc. Annual Meeting of
  the Association for Computational Linguistics} (2015)

\bibitem{BritzGoldieEtAl2017}
Britz, D., Goldie, A., Luong, M.T., Le, Q.V.: Massive exploration of neural
  machine translation architectures. In: {Proc.\ of Conf.\ on Empirical Methods
  in Natural Language Processing}. pp. 1442--1451 (2017)

\bibitem{CaesarBankitiEtAl2019}
Caesar, H., Bankiti, V., Lang, A.H., Vora, S., Liong, V.E., Xu, Q., Krishnan,
  A., Pan, Y., Baldan, G., Beijbom, O.: {nuScenes}: A multimodal dataset for
  autonomous driving (2019)

\bibitem{CasasGulinoEtAl2019}
Casas, S., Gulino, C., Liao, R., Urtasun, R.: {SpAGNN}: Spatially-aware graph
  neural networks for relational behavior forecasting from sensor data (2019)

\bibitem{CasasLuoEtAl2018}
Casas, S., Luo, W., Urtasun, R.: {IntentNet}: Learning to predict intention
  from raw sensor data. In: {Conf.\ on Robot Learning}. pp. 947--956 (2018)

\bibitem{ChangLambertEtAl2019}
Chang, M.F., Lambert, J., Sangkloy, P., Singh, J., Bak, S., Hartnett, A., Wang,
  D., Carr, P., Lucey, S., Ramanan, D., Hays, J.: Argoverse: 3d tracking and
  forecasting with rich maps. In: {IEEE Conf.\ on Computer Vision and Pattern
  Recognition} (2019)

\bibitem{ChoMerrienboerEtAl2014}
Cho, K., van Merrienboer, B., Gulcehre, C., Bahdanau, D., Bougares, F.,
  Schwenk, H., Bengio, Y.: Learning phrase representations using rnn
  encoder-decoder for statistical machine translation. In: {Proc.\ of Conf.\ on
  Empirical Methods in Natural Language Processing}. pp. 1724--1734 (2014)

\bibitem{DeoTrivedi2018}
Deo, N., Trivedi, M.M.: Multi-modal trajectory prediction of surrounding
  vehicles with maneuver based lstms. In: {IEEE Intelligent Vehicles Symposium}
  (2018)

\bibitem{GoodfellowPouget-AbadieEtAl2014}
Goodfellow, I., Pouget-Abadie, J., Mirza, M., Xu, B., Warde-Farley, D., Ozair,
  S., Courville, A., Bengio, Y.: Generative adversarial nets. In: {Conf.\ on
  Neural Information Processing Systems} (2014)

\bibitem{GuptaJohnsonEtAl2018}
Gupta, A., Johnson, J., Li, F., Savarese, S., Alahi, A.: Social {GAN}: Socially
  acceptable trajectories with generative adversarial networks. In: {IEEE
  Conf.\ on Computer Vision and Pattern Recognition} (2018)

\bibitem{GweonSaxe2013}
Gweon, H., Saxe, R.: Developmental cognitive neuroscience of theory of mind.
  In: Neural Circuit Development and Function in the Brain, chap.~20, pp.
  367--377. {Academic Press} (2013).
  \doi{https://doi.org/10.1016/B978-0-12-397267-5.00057-1},
  \url{http://www.sciencedirect.com/science/article/pii/B9780123972675000571}

\bibitem{HallacLeskovecEtAl2015}
Hallac, D., Leskovec, J., Boyd, S.: Network lasso: Clustering and optimization
  in large graphs. In: {ACM Int.\ Conf.\ on Knowledge Discovery and Data
  Mining} (2015)

\bibitem{HelbingMolnar1995}
Helbing, D., Moln\'{a}r, P.: Social force model for pedestrian dynamics.
  {Physical Review E}  \textbf{51}(5),  4282--4286 (1995)

\bibitem{HigginsMattheyEtAl2017}
Higgins, I., Matthey, L., Pal, A., Burgess, C., Glorot, X., Botvinick, M.,
  Mohamed, S., Lerchner, A.: {beta-VAE}: Learning basic visual concepts with a
  constrained variational framework. In: {Int.\ Conf.\ on Learning
  Representations} (2017)

\bibitem{HochreiterSchmidhuber1997}
Hochreiter, S., Schmidhuber, J.: Long short-term memory. {Neural Computation}
  (1997)

\bibitem{IvanovicPavone2019}
Ivanovic, B., Pavone, M.: The {Trajectron}: Probabilistic multi-agent
  trajectory modeling with dynamic spatiotemporal graphs. In: {IEEE Int.\
  Conf.\ on Computer Vision} (2019)

\bibitem{IvanovicSchmerlingEtAl2018}
Ivanovic, B., Schmerling, E., Leung, K., Pavone, M.: Generative modeling of
  multimodal multi-human behavior. In: {IEEE/RSJ Int.\ Conf.\ on Intelligent
  Robots \& Systems} (2018)

\bibitem{JainZamirEtAl2016}
Jain, A., Zamir, A.R., Savarese, S., Saxena, A.: Structural-{RNN}: Deep
  learning on spatio-temporal graphs. In: {IEEE Conf.\ on Computer Vision and
  Pattern Recognition} (2016)

\bibitem{JainCasasEtAl2019}
Jain, A., Casas, S., Liao, R., Xiong, Y., Feng, S., Segal, S., Urtasun, R.:
  Discrete residual flow for probabilistic pedestrian behavior prediction. In:
  {Conf.\ on Robot Learning} (2019)

\bibitem{JangGuEtAl2017}
Jang, E., Gu, S., Poole, B.: Categorial reparameterization with gumbel-softmax.
  In: {Int.\ Conf.\ on Learning Representations} (2017)

\bibitem{Kalman1960}
Kalman, R.E.: A new approach to linear filtering and prediction problems. {ASME
  Journal of Basic Engineering}  \textbf{82},  35--45 (1960)

\bibitem{lyft_dataset2019}
Kesten, R., Usman, M., Houston, J., Pandya, T., Nadhamuni, K., Ferreira, A.,
  Yuan, M., Low, B., Jain, A., Ondruska, P., Omari, S., Shah, S., Kulkarni, A.,
  Kazakova, A., Tao, C., Platinsky, L., Jiang, W., Shet, V.: {Lyft Level 5 AV
  Dataset 2019}. \url{https://level5.lyft.com/dataset/} (2019)

\bibitem{KongPfeiferEtAl2015}
Kong, J., Pfeifer, M., Schildbach, G., Borrelli, F.: Kinematic and dynamic
  vehicle models for autonomous driving control design. In: {IEEE Intelligent
  Vehicles Symposium} (2015)

\bibitem{KosarajuSadeghianEtAl2019}
Kosaraju, V., Sadeghian, A., Mart\'{i}n-Mart\'{i}n, R., Reid, I., Rezatofighi,
  S.H., Savarese, S.: {Social-BiGAT}: Multimodal trajectory forecasting using
  bicycle-{GAN} and graph attention networks. In: {Conf.\ on Neural Information
  Processing Systems} (2019)

\bibitem{LaValle2006BetterUnicycle}
LaValle, S.M.: Better unicycle models. In: Planning Algorithms, pp. 743--743.
  {Cambridge Univ.\ Press} (2006)

\bibitem{LaValle2006Unicycle}
LaValle, S.M.: A simple unicycle. In: Planning Algorithms, pp. 729--730.
  {Cambridge Univ.\ Press} (2006)

\bibitem{LeeChoiEtAl2017}
Lee, N., Choi, W., Vernaza, P., Choy, C.B., Torr, P.H.S., Chandraker, M.:
  {DESIRE:} distant future prediction in dynamic scenes with interacting
  agents. In: {IEEE Conf.\ on Computer Vision and Pattern Recognition} (2017)

\bibitem{LeeKitani2016}
Lee, N., Kitani, K.M.: Predicting wide receiver trajectories in {A}merican
  football. In: {IEEE Winter Conf.\ on Applications of Computer Vision} (2016)

\bibitem{LernerChrysanthouEtAl2007}
Lerner, A., Chrysanthou, Y., Lischinski, D.: Crowds by example. {Computer
  Graphics Forum}  \textbf{26}(3),  655--664 (2007)

\bibitem{MortonWheelerEtAl2017}
Morton, J., Wheeler, T.A., Kochenderfer, M.J.: Analysis of recurrent neural
  networks for probabilistic modeling of driver behavior. {IEEE Transactions on
  Pattern Analysis \& Machine Intelligence}  \textbf{18}(5),  1289--1298 (2017)

\bibitem{PadenCapEtAl2016}
Paden, B., \v{C}\'{a}p, M., Yong, S.Z., Yershov, D., Frazzoli, E.: A survey of
  motion planning and control techniques for self-driving urban vehicles. {IEEE
  Transactions on Intelligent Vehicles}  \textbf{1}(1),  33--55 (2016)

\bibitem{PaszkeGrossEtAl2017}
Paszke, A., Gross, S., Chintala, S., Chanan, G., Yang, E., DeVito, Z., Lin, Z.,
  Desmaison, A., Antiga, L., Lerer, A.: Automatic differentiation in {PyTorch}.
  In: {Conf.\ on Neural Information Processing Systems - Autodiff Workshop}
  (2017)

\bibitem{PellegriniEssEtAl2009}
Pellegrini, S., Ess, A., Schindler, K., Gool, L.v.: You'll never walk alone:
  Modeling social behavior for multi-target tracking. In: {IEEE Int.\ Conf.\ on
  Computer Vision} (2009)

\bibitem{RasmussenWilliams2006}
Rasmussen, C.E., Williams, C.K.I.: Gaussian Processes for Machine Learning
  (Adaptive Computation and Machine Learning). {MIT Press}, first edn. (2006)

\bibitem{RhinehartMcAllisterEtAl2019}
Rhinehart, N., McAllister, R., Kitani, K., Levine, S.: {PRECOG}: Prediction
  conditioned on goals in visual multi-agent settings. In: {IEEE Int.\ Conf.\
  on Computer Vision} (2019)

\bibitem{RudenkoPalmieriEtAl2019}
Rudenko, A., Palmieri, L., Herman, M., Kitani, K.M., Gavrila, D.M., Arras,
  K.O.: Human motion trajectory prediction: A survey. {Int.\ Journal of
  Robotics Research}  \textbf{39}(8),  895--935 (2020)

\bibitem{SadeghianKosarajuEtAl2019}
Sadeghian, A., Kosaraju, V., Sadeghian, A., Hirose, N., Rezatofighi, S.H.,
  Savarese, S.: {SoPhie}: An attentive {GAN} for predicting paths compliant to
  social and physical constraints. In: {IEEE Conf.\ on Computer Vision and
  Pattern Recognition} (2019)

\bibitem{SadeghianLegrosEtAl2018}
Sadeghian, A., Legros, F., Voisin, M., Vesel, R., Alahi, A., Savarese, S.:
  {CAR-Net}: Clairvoyant attentive recurrent network. In: {European Conf.\ on
  Computer Vision} (2018)

\bibitem{SchollerAravantinosEtAl2020}
Sch\"{o}ller, C., Aravantinos, V., Lay, F., Knoll, A.: What the constant
  velocity model can teach us about pedestrian motion prediction. {IEEE
  Robotics and Automation Letters}  (2020)

\bibitem{SohnLeeEtAl2015}
Sohn, K., Lee, H., Yan, X.: Learning structured output representation using
  deep conditional generative models. In: {Conf.\ on Neural Information
  Processing Systems} (2015)

\bibitem{TangSalakhutdinov2019}
Tang, Y.C., Salakhutdinov, R.: Multiple futures prediction. In: {Conf.\ on
  Neural Information Processing Systems} (2019)

\bibitem{ThiedeBrahma2019}
Thiede, L.A., Brahma, P.P.: Analyzing the variety loss in the context of
  probabilistic trajectory prediction. In: {IEEE Int.\ Conf.\ on Computer
  Vision} (2019)

\bibitem{ThrunBurgardEtAl2005EKF}
Thrun, S., Burgard, W., Fox, D.: The extended {Kalman} filter. In:
  Probabilistic Robotics, pp. 54--64. {MIT Press} (2005)

\bibitem{VemulaMuellingEtAl2018}
Vemula, A., Muelling, K., Oh, J.: Social attention: Modeling attention in human
  crowds. In: {Proc.\ IEEE Conf.\ on Robotics and Automation} (2018)

\bibitem{WangFleetEtAl2008}
Wang, J.M., Fleet, D.J., Hertzmann, A.: Gaussian process dynamical models for
  human motion. {IEEE Transactions on Pattern Analysis \& Machine Intelligence}
   \textbf{30}(2),  283--298 (2008)

\bibitem{WaymoSafety2018}
Waymo: Safety report (2018), {Available at }\url{https://waymo.com/safety/}.
  Retrieved on November 9, 2019

\bibitem{waymo_open_dataset}
Waymo: {Waymo Open Dataset}: An autonomous driving dataset.
  \url{https://waymo.com/open/} (2019)

\bibitem{ZengLuoEtAl2019}
Zeng, W., Luo, W., Suo, S., Sadat, A., Yang, B., Casas, S., Urtasun, R.:
  End-to-end interpretable neural motion planner. In: {IEEE Conf.\ on Computer
  Vision and Pattern Recognition} (2019)

\bibitem{ZhaoSongEtAl2019}
Zhao, S., Song, J., Ermon, S.: {InfoVAE}: Balancing learning and inference in
  variational autoencoders. In: {Proc.\ AAAI Conf.\ on Artificial Intelligence}
  (2019)

\bibitem{ZhaoXuEtAl2019}
Zhao, T., Xu, Y., Monfort, M., Choi, W., Baker, C., Zhao, Y., Wang, Y., Wu,
  Y.N.: Multi-agent tensor fusion for contextual trajectory prediction. In:
  {IEEE Conf.\ on Computer Vision and Pattern Recognition} (2019)

\end{thebibliography}

\appendix

\section{Single Integrator Distribution Integration} \label{sec:si_integration}
For a single integrator, we define the state to be the position vector $\mathbf{s} = \mathbf{p} = [x, y]^T$, the control to be the velocity vector $\mathbf{u} = \dot{\mathbf{p}} = [\dot{x}, \dot{y}]^T$, and write the linear discrete-time dynamics as
\begin{equation}\label{eqn:supp_SI}
\mathbf{p}^{(t+1)} = I_{2 \times 2} \mathbf{p}^{(t)} + \Delta t I_{2 \times 2} \dot{\mathbf{p}}^{(t)}.
\end{equation}
At each timestep, and for a specific latent value $z$, \emphalgname{} produces a Gaussian distribution over control actions $\mathcal{N}(\mu_{\mathbf{u}}, \mathbf{\Sigma}_{\mathbf{u}})$. Specifically, it outputs
\begin{equation}
\mu_{\mathbf{u}} = \begin{bmatrix} 
\mu_{\dot{x}}\\
\mu_{\dot{y}}
\end{bmatrix} \hspace{1cm} \mathbf{\Sigma}_{\mathbf{u}} = \begin{bmatrix} 
\sigma_{\dot{x}}^2 & \rho_{\dot{x}\dot{y}} \sigma_{\dot{x}} \sigma_{\dot{y}}\\
\rho_{\dot{x}\dot{y}} \sigma_{\dot{x}} \sigma_{\dot{y}} & \sigma_{\dot{y}}^2
\end{bmatrix},
\end{equation}
where $\mu_{\dot{x}}$ and $\mu_{\dot{y}}$ are the respective mean velocities in the agent's longitudinal and lateral directions, $\sigma_{\dot{x}}$ and $\sigma_{\dot{y}}$ are the respective longitudinal and lateral velocity standard deviations, and $\rho_{\dot{x}\dot{y}}$ is the correlation between $\dot{x}$ and $\dot{y}$. Since $\mathbf{\Sigma}_{\mathbf{u}}$ is the only source of uncertainty in the prediction model, \cref{eqn:supp_SI} is a linear Gaussian system.

\subsection{Mean Derivation\protect\footnote{These equations are also found in the Kalman Filter prediction step~\cite{Kalman1960}.}}
Following the sum of Gaussian random variables~\cite{Kalman1960}, the output mean positions are obtained by \cref{eqn:supp_SI}. Thus, at test time, \emphalgname{} produces $\mu_{\dot{\mathbf{p}}}^{(t)}$ which is passed through \cref{eqn:supp_SI} alongside the current agent position $\mu_{\mathbf{p}}^{(t)}$ to produce the predicted position mean $\mu_{\mathbf{p}}^{(t+1)}$.

\subsection{Covariance Derivation\protect\footnotemark[2]}
The position covariance is obtained via the covariance of a sum of Gaussian random variables~\cite{Kalman1960}
\begin{equation}\label{eqn:supp_SI_cov_update}
\begin{aligned}
\mathbf{\Sigma}_{\mathbf{p}}^{(t+1)} &= I_{2 \times 2} \mathbf{\Sigma}_{\mathbf{p}}^{(t)} I_{2 \times 2}^T + \Delta t I_{2 \times 2} \mathbf{\Sigma}_{\mathbf{u}}^{(t)} \Delta t I_{2 \times 2}^T\\
&= \mathbf{\Sigma}_{\mathbf{p}}^{(t)} + (\Delta t)^2 \mathbf{\Sigma}_{\mathbf{u}}^{(t)}.
\end{aligned}
\end{equation}

\section{Dynamically-Extended Unicycle Distribution Integration}\label{sec:unicycle_integration}
Usually, unicycle models have velocity and heading rate as control inputs \cite{LaValle2006Unicycle,PadenCapEtAl2016}. However, vehicles in the real world are controlled by accelerator pedals and so we instead adopt the dynamically-extended unicycle model which instead uses acceleration $a$ and heading rate $\omega$ as control inputs \cite{LaValle2006BetterUnicycle}. The dynamically-extended unicycle model has the following nonlinear continuous-time dynamics
\begin{equation}\label{eqn:supp_unicycle_continuous}
\begin{bmatrix}
\dot{x}\\
\dot{y}\\
\dot{\phi}\\
\dot{v}
\end{bmatrix} = \begin{bmatrix}
v \cos (\phi)\\
v \sin (\phi)\\
\omega\\
a
\end{bmatrix},
\end{equation}
where $\mathbf{p} = [x, y]^T$ defines the position, $v$ the speed, and $\phi$ the heading. As mentioned above, the control inputs are $\mathbf{u} = [\omega, a]^T$. To discretize this, we assume a zero-order hold on the controls between each sampling step (i.e. control actions are piece-wise constant). This yields the following zero-order hold discrete equivalent dynamics
\begin{equation}\label{eqn:supp_unicycle}
\begin{aligned}
\begin{bmatrix}
x^{(t+1)}\\
y^{(t+1)}\\
\phi^{(t+1)}\\
v^{(t+1)}
\end{bmatrix} &= \begin{bmatrix}
x^{(t)}\\
y^{(t)}\\
\phi^{(t)}\\
v^{(t)}
\end{bmatrix} + \begin{bmatrix}
v^{(t)} \cdot D_S^{(t)} + \frac{a^{(t)} \sin(\phi^{(t)} + \omega^{(t)} \Delta t) \Delta t}{\omega^{(t)}} + \frac{a^{(t)}}{\omega^{(t)}} \cdot D_C^{(t)}\\
- v^{(t)} \cdot D_C^{(t)} - \frac{a^{(t)} \cos(\phi^{(t)} + \omega^{(t)} \Delta t) \Delta t }{\omega^{(t)}} + \frac{a^{(t)}}{\omega^{(t)}} \cdot D_S^{(t)}\\
\omega^{(t)} \Delta t\\
a^{(t)} \Delta t
\end{bmatrix},\\
\text{where } D_S^{(t)} &= \frac{\sin(\phi^{(t)} + \omega^{(t)} \Delta t) - \sin(\phi^{(t)})}{\omega^{(t)}}\\
D_C^{(t)} &= \frac{\cos(\phi^{(t)} + \omega^{(t)} \Delta t) - \cos(\phi^{(t)})}{\omega^{(t)}}.
\end{aligned}
\end{equation}
We will refer to these dynamics in short with $\mathbf{s}^{(t+1)} = \mathbf{f}(\mathbf{s}^{(t)}, \mathbf{u}^{(t)})$. We adopt a slightly different set of dynamics when $|\omega| \leq \epsilon = 10^{-3}$ to avoid singularities in \cref{eqn:supp_unicycle}. With a small $\omega$, we instead use the following dynamics, obtained by evaluating the limit as $\omega \rightarrow 0$.
\begin{equation}\label{eqn:supp_unicycle_small_omega}
\begin{aligned}
\begin{bmatrix}
x^{(t+1)}\\
y^{(t+1)}\\
\phi^{(t+1)}\\
v^{(t+1)}
\end{bmatrix} &= \begin{bmatrix}
x^{(t)}\\
y^{(t)}\\
\phi^{(t)}\\
v^{(t)}
\end{bmatrix} + \begin{bmatrix}
v^{(t)} \cos(\phi^{(t)}) \Delta t + 0.5 a^{(t)} \cos(\phi^{(t)}) (\Delta t)^2\\
v^{(t)} \sin(\phi^{(t)}) \Delta t + 0.5 a^{(t)} \sin(\phi^{(t)}) (\Delta t)^2\\
0\\
a^{(t)} \Delta t
\end{bmatrix}.
\end{aligned}
\end{equation}
Thus, the full discrete-time dynamics are
\begin{equation}\label{eqn:supp_unicycle_full}
\begin{bmatrix}
x^{(t+1)}\\
y^{(t+1)}\\
\phi^{(t+1)}\\
v^{(t+1)}
\end{bmatrix} = \begin{cases}
\text{\cref{eqn:supp_unicycle}} & \text{if } |\omega| > \epsilon\\
\text{\cref{eqn:supp_unicycle_small_omega}} & \text{otherwise}
\end{cases}.
\end{equation}

At each timestep, and for a specific latent value $z$, \emphalgname{} produces a Gaussian distribution over control actions $\mathcal{N}(\mu_{\mathbf{u}}, \mathbf{\Sigma}_{\mathbf{u}})$. Specifically, it outputs
\begin{equation}
\mu_{\mathbf{u}} = \begin{bmatrix}
\mu_{\omega}\\
\mu_{a}
\end{bmatrix} \hspace{1cm} \mathbf{\Sigma}_{\mathbf{u}} = \begin{bmatrix} 
\sigma_{\omega}^2 & \rho_{\omega a} \sigma_{\omega} \sigma_{a}\\
\rho_{\omega a} \sigma_{\omega} \sigma_{a} & \sigma_{a}^2
\end{bmatrix},
\end{equation}
where $\mu_{\omega}$ is the mean rate of change of the agent's heading, $\mu_{a}$ is the mean acceleration in the agent's heading direction, $\sigma_{\omega}$ is the standard deviation of the heading rate of change, $\sigma_{a}$ is the acceleration standard deviation, and $\rho_{\omega a}$ is the correlation between $\omega$ and $a$. The controls $\mu_{\mathbf{u}}$ and uncertainties $\mathbf{\Sigma}_{\mathbf{u}}$ are then integrated through the dynamics to obtain the following mean and covariance integration equations~\cite{ThrunBurgardEtAl2005EKF}.


\subsection{Mean Derivation\protect\footnote{These equations are also found in the Extended Kalman Filter prediction step~\cite{ThrunBurgardEtAl2005EKF}.}}
The output mean positions are obtained by applying the mean control actions to \cref{eqn:supp_unicycle_full}~\cite{ThrunBurgardEtAl2005EKF}.

\subsection{Covariance Derivation\protect\footnotemark[3]}
Since $\mathbf{\Sigma}_{\mathbf{u}}$ is the only source of uncertainty in the prediction model, \cref{eqn:supp_unicycle_full} can be made a linear Gaussian system by linearizing about a specific state and control. For instance, the Jacobians $\mathbf{F}$ and $\mathbf{G}$ of the system dynamics in \cref{eqn:supp_unicycle} are
\begin{equation}
\begin{aligned}
\mathbf{F}^{(t)} &= \frac{\partial \mathbf{f}}{\partial \mu_{\mathbf{s}}^{(t)}} = \begin{bmatrix}
1 & 0 & v^{(t)} D_C^{(t)} - \frac{a^{(t)} D_S^{(t)}}{\omega^{(t)}} + \frac{a^{(t)} \cos(\phi^{(t)} + \omega^{(t)} \Delta t) \Delta t}{\omega^{(t)}} & D_S^{(t)}\\
0 & 1 & v^{(t)} D_S^{(t)} + \frac{a^{(t)} D_C^{(t)}}{\omega^{(t)}} + \frac{a^{(t)} \sin(\phi^{(t)} + \omega^{(t)} \Delta t) \Delta t}{\omega^{(t)}} & - D_C^{(t)}\\
0 & 0 & 1 & 0\\
0 & 0 & 0 & 1
\end{bmatrix}\\
\mathbf{G}^{(t)} &= \frac{\partial \mathbf{f}}{\partial \mu_{\mathbf{u}}^{(t)}} = \begin{bmatrix}
G_{11}^{(t)} & \frac{D_C^{(t)}}{\omega^{(t)}} + \frac{\sin(\phi^{(t)} + \omega^{(t)} \Delta t) \Delta t}{\omega^{(t)}}\\
G_{21}^{(t)} & \frac{D_S^{(t)}}{\omega^{(t)}} - \frac{\cos(\phi^{(t)} + \omega^{(t)} \Delta t) \Delta t}{\omega^{(t)}}\\
\Delta t & 0\\
0 & \Delta t
\end{bmatrix},\\
\text{where } G_{11}^{(t)} &= \frac{v \cos(\phi + \omega \Delta t) \Delta t}{\omega} - \frac{v D_S}{\omega} - \frac{2a \sin(\phi + \omega \Delta t) \Delta t}{\omega^2} - \frac{2 a D_C}{\omega^2} \\
&\hspace{0.5cm}+ \frac{a \cos(\phi + \omega \Delta t) (\Delta t)^2}{\omega}\\
G_{21}^{(t)} &= \frac{v \sin(\phi + \omega \Delta t) \Delta t}{\omega} + \frac{v D_C}{\omega} + \frac{2a \cos(\phi + \omega \Delta t) \Delta t}{\omega^2} - \frac{2 a D_S}{\omega^2}\\
&\hspace{0.5cm} + \frac{a \sin(\phi + \omega \Delta t) (\Delta t)^2}{\omega}.
\end{aligned}
\end{equation}
Then, applying the equations for the covariance of a sum of Gaussian random variables~\cite{ThrunBurgardEtAl2005EKF} yields
\begin{equation}\label{eqn:supp_unicycle_cov_update}
\mathbf{\Sigma}_{\mathbf{p}, \theta, v}^{(t+1)} = \mathbf{F}^{(t)} \mathbf{\Sigma}_{\mathbf{p}, \theta, v}^{(t)} \mathbf{F}^{{(t)}^T} + \mathbf{G}^{(t)} \mathbf{\Sigma}_{\mathbf{u}}^{(t)} \mathbf{G}^{{(t)}^T}.
\end{equation}

\section{Average and Final Displacement Error Evaluation} \label{sec:supp_ADE}

\begin{figure}[t]
    \centering
    \includegraphics[width=0.495\linewidth]{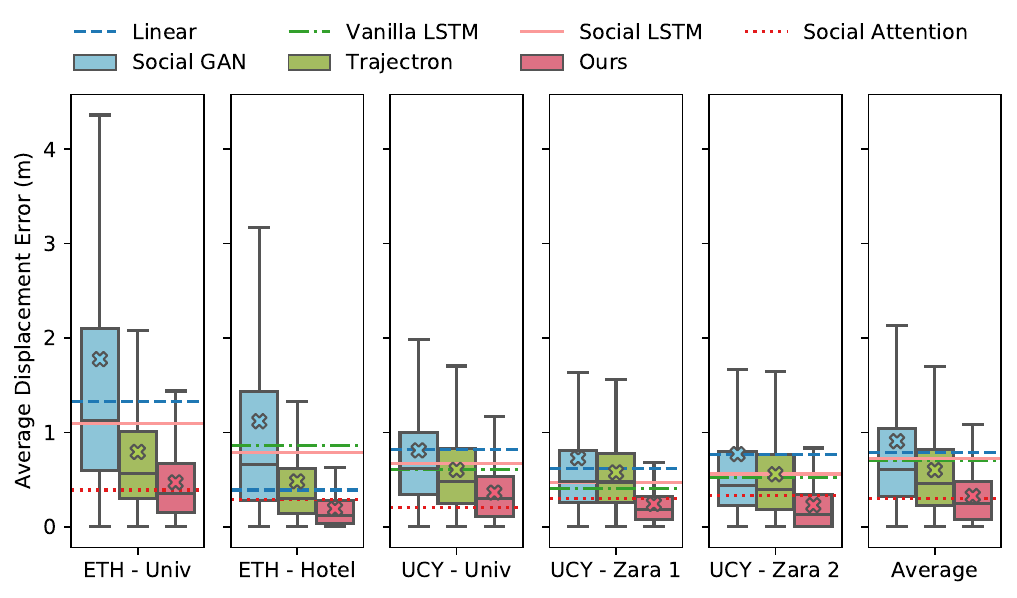}
    \includegraphics[width=0.495\linewidth]{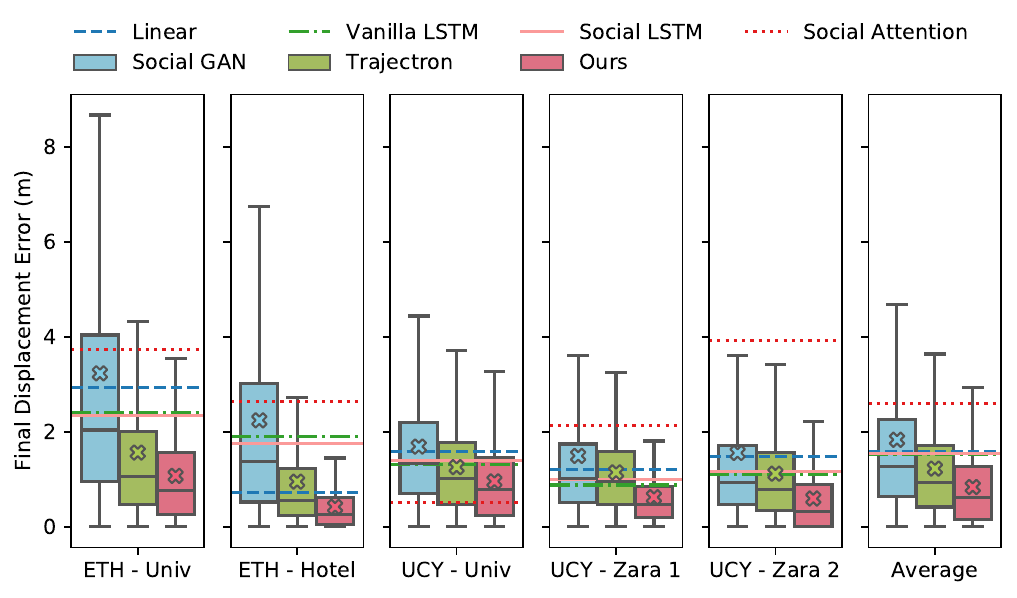}
    \caption{\textbf{Left}: ADE results of all methods per dataset, as well as their average performance. Boxplots are shown for all generative models since they produce distributions of trajectories. 2000 trajectories were sampled per model at each prediction timestep, with each sample’s ADE included in the boxplots. Our approach with dynamics integration is compared here, specifically its $z_\text{mode}$ output configuration. X markers indicate the mean ADE. Mean ADE from deterministic baselines are visualized as horizontal lines. \textbf{Right}: The same analysis for FDE.}
    \label{fig:supp_ADE}
    \label{fig:supp_FDE}
    \vspace{3mm}
\end{figure}

While ADE and FDE are important metrics for deterministic, single-trajectory methods, any deeper probabilistic information available from generative methods is destroyed when taking the mean over the dataset. Instead, in the main body of the paper we focus on evaluation methods which maintain such information. However, we can somewhat directly compare deterministic and generative methods using ADE and FDE by directly plotting the full error distributions for any generative methods, as in \cite{IvanovicPavone2019}. This provides an idea as to how close and concentrated the predictions are around the ground truth. \cref{fig:supp_ADE} shows both generative and deterministic methods' ADE and FDE performance. In both metrics, our method's error distribution is lower and more concentrated than other generative approaches, even outperforming state-of-the-art deterministic methods. 




\section{Additional Training Information}

\subsection{Choosing $\alpha, \beta$ in \cref{eqn:loss_fn}}

As shown in \cite{HigginsMattheyEtAl2017}, the $\beta$ parameter weighting the KL penalty term is important to disentangle the latent space and encourage multimodality. A good value for this hyperparameter varies with the the size of input $\mathbf{y}$, condition $\mathbf{x}$, and latent space $z$. Therefore, we adjust $\beta$ depending on the size of the encoder's output $e_\mathbf{x}$. For example, we increase the value of $\beta$ when encoding map information in the condition. Additionally, $\beta$ is annealed following an increasing sigmoid \cite{BowmanVilnisEtAl2015}. Thus, a low $\beta$ factor is used during early training iterations so that the model learns to encode as much information in $z$ as possible. As training continues, $\beta$ is gradually increased to shift the role of information encoding from $q_\phi(z \mid \mathbf{x}, \mathbf{y})$ to $p_\theta(z \mid \mathbf{x})$. For $\alpha$, we found that a constant value of $1.0$ works well.

\subsection{Separate Map Encoder Learning Rate}
When used, we train the map encoding CNN with a smaller learning rate compared to the rest of the model, to prevent large gradients in early training iterations. We use leaky ReLU activation functions with $\alpha=0.2$ to prevent saturation during early training iterations (when the CNN does not provide useful encodings to the rest of the model). We found that regular ReLU, sigmoid, and tanh activation functions saturate during early training and fail to recover.

\subsection{Data Augmentation}

To avoid overfitting to environment-specific characteristics, such as the general directions that agents move, we augment the data from each scene. We rotate all trajectories in a scene around the scene's origin by $\gamma$, where $\gamma$ varies from $0\degree$ to $360\degree$ (exclusive) in $15\degree$ intervals. The benefits of dataset augmentation by trajectory rotation have already been studied for pedestrians \cite{SchollerAravantinosEtAl2020}. We apply this same augmentation to autonomous driving datasets as most of them are recorded in cities whose streets are roughly orthogonal and separated by blocks.

\section{ETH Pedestrians Map Encoding}

\begin{figure}[t]
    \centering
    \includegraphics[height=2.6cm,frame]{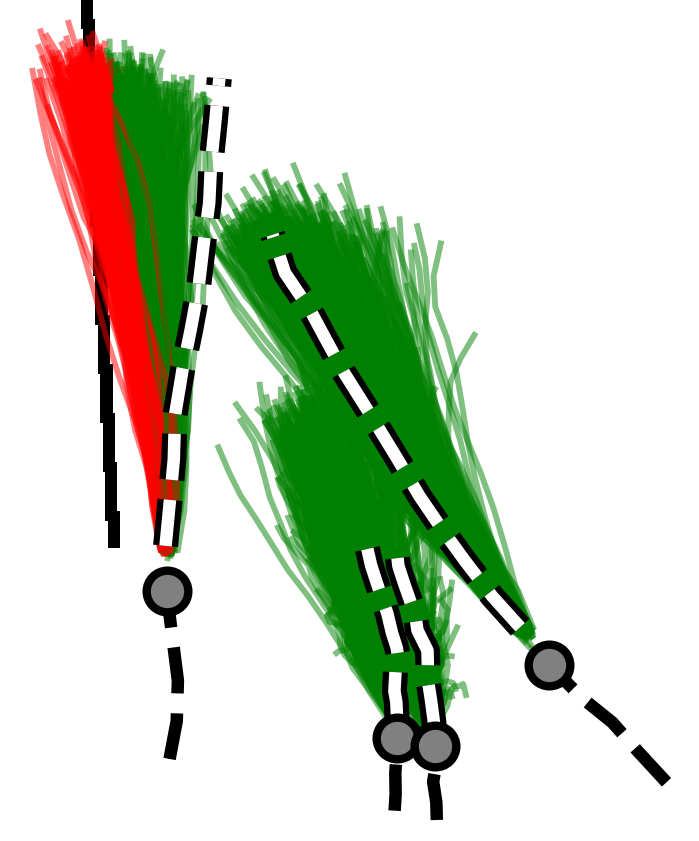}
    \includegraphics[height=2.6cm,frame]{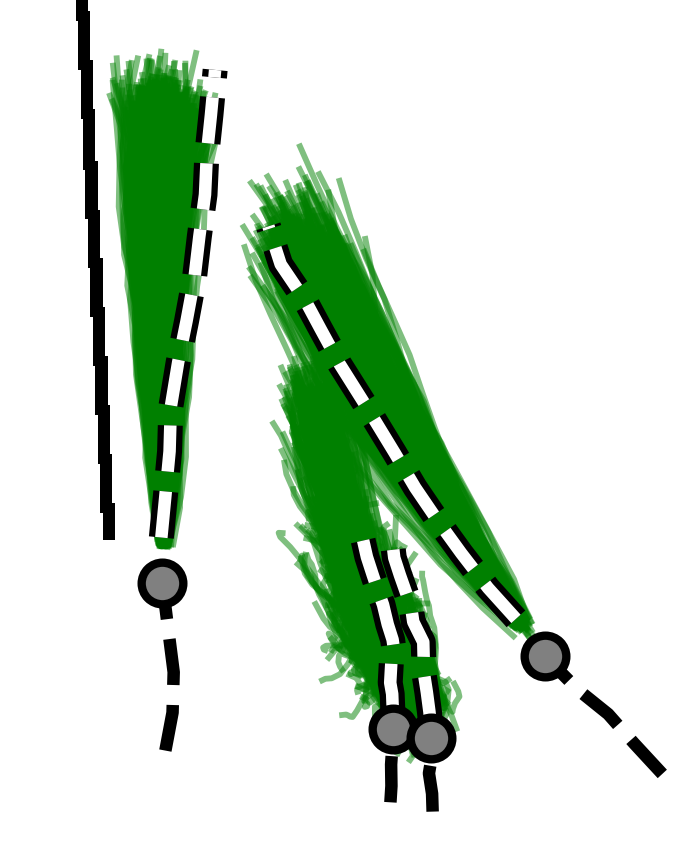}
    \raisebox{0.65cm}{\includegraphics[height=1.25cm]{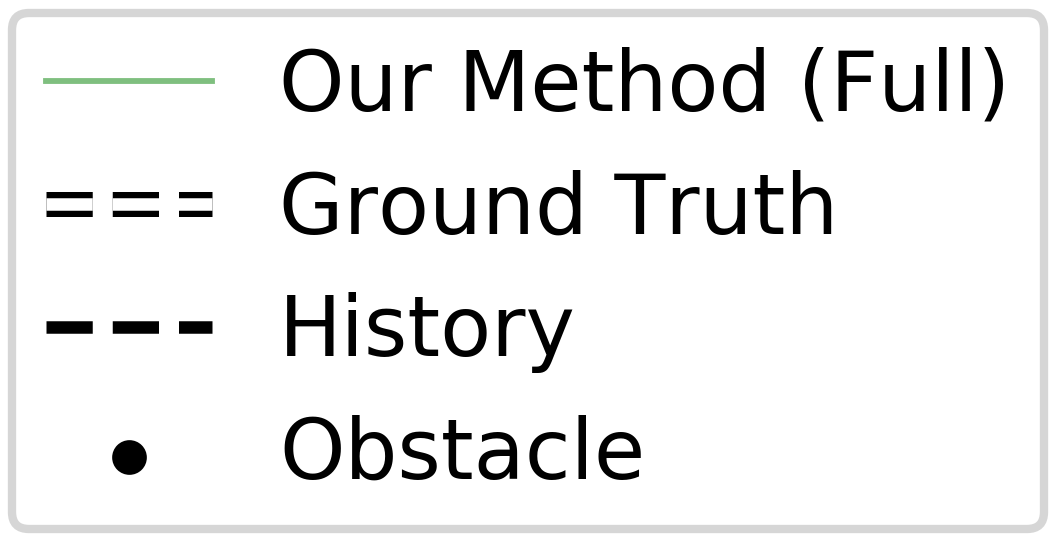}}
    \caption{\textbf{Left:} When only using trajectory data, \emphalgname{} does not know of obstacles and makes predictions into walls (in red). \textbf{Right:} Encoding a local map of the agent's surroundings significantly reduces the frequency of obstacle-violating predictions.}
    \label{fig:supp_eth_map_quali}
\end{figure}

\cref{fig:supp_eth_map_quali} shows an example of the reduction in the number of obstacle violations for pedestrians in the ETH - University scene that are especially close to an obstacle (i.e. they have at least one obstacle-violating trajectory in their Full output).

\section{Online Runtime}

\begin{figure}[t]
    \centering
    \includegraphics[width=\linewidth]{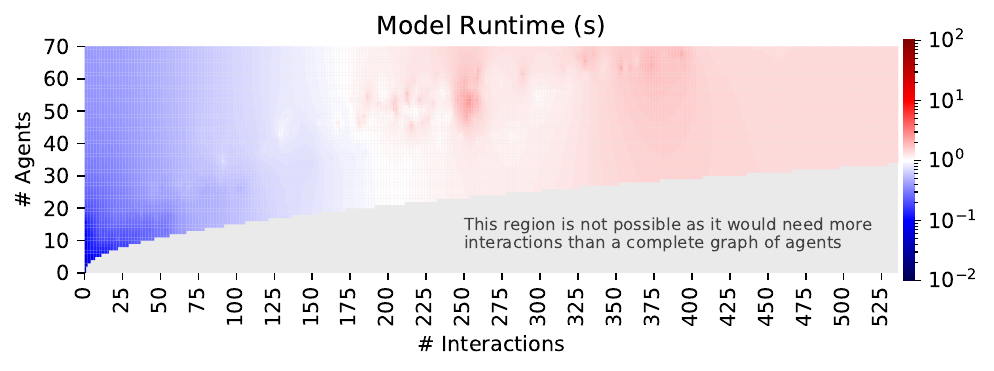}
    \caption{Mean time for \emphalgname{} to generate future trajectory distributions on a laptop with a 2.7 GHz Intel Core i5 (Broadwell) CPU and 8 GB of RAM. None of the runtimes exceed $1.2s$ (with most at or below $1s$), enabling the model to run $3 - 5\times$ per prediction horizon for all tested datasets.}
    \label{fig:supp_runtime}
    \vspace{3mm}
\end{figure}

\cref{fig:supp_runtime} illustrates how \emphalgname{}'s runtime scales with respect to problem size. In particular, a heatmap is used as there are two major factors that affect the model's runtime: number of agents and amount of interactions. For points with insufficient data, e.g., the rare case of 50 agents in a scene with only 3 interactions, we impute values using an optimization-based scheme~\cite{HallacLeskovecEtAl2015}.
To achieve this real-time performance, we leverage the stateful representation that spatiotemporal graphs provide. Specifically, \emphalgname{} is updated online with new information without fully executing a forward pass. This is possible due to our method's use of LSTMs, as only the last LSTM cells in the encoder need to be fed the newly-observed data. The rest of the model can then be executed using the updated encoder representation.

\end{document}